\documentclass{article}
\usepackage{tencent_digital_human}
\usepackage[colorlinks = true,
            linkcolor = blue,
            urlcolor  = blue,
            citecolor = blue,
            anchorcolor = blue]{hyperref}

\usepackage[most]{tcolorbox} 
\definecolor{tsinghuapurple}{RGB}{102,8,116}
\newtcolorbox{alprompt}[1]{
        boxrule = 1pt,
        fontupper = \small\tt,
        fonttitle = \bf\color{black},
        arc = 2pt,
        rounded corners,
        colframe = black,
        colbacktitle = white!97!yellow,
        colback = white!97!yellow,
        title = #1,
}

\usepackage{amssymb}
\usepackage{microtype}
\usepackage{hyperref}
\usepackage{url}
\usepackage{booktabs}
\usepackage{enumitem}
\usepackage{multicol}
\usepackage{multirow}
\usepackage{CJKutf8}
\usepackage{amsmath}
\usepackage{siunitx}
\usepackage{floatflt}
\usepackage{wrapfig}
\usepackage{graphicx}
\usepackage{booktabs}
\usepackage{wrapfig}
\usepackage{authblk}
\usepackage{lipsum}

\usepackage{algorithm}
\usepackage{algorithmicx}
\usepackage{algpseudocode}
\usepackage{microtype}
\usepackage{graphicx}
\usepackage{multirow}
\usepackage{booktabs} 
\usepackage{pifont}  
\usepackage{graphicx}  
\usepackage{subcaption} 
\usepackage{hyperref}

\usepackage{amssymb} 

\algnewcommand{\LeftComment}[1]{\Statex \(\triangleright\) #1}

\newtcolorbox{promptbox}[3][Prompt]{
colback=black!5!white,
arc=5pt, 
boxrule=0.5pt,
fonttitle=\bfseries,
title=#1, 
before upper={\small}, fontupper=\fontfamily{ptm}\selectfont,
colframe=#2,
label=#3,
}
\usepackage{array}
\usepackage{amsmath}
\usepackage{amssymb}
\usepackage{mathtools}
\usepackage{amsthm}
\usepackage{arydshln}
\usepackage[capitalize,noabbrev]{cleveref}
\usepackage{adjustbox} 
\usepackage{enumitem}
\usepackage{xspace}
\theoremstyle{plain}

\theoremstyle{definition}

\theoremstyle{remark}

\usepackage{xcolor}
\tcbuselibrary{listings,skins}

\definecolor{promptbg}{RGB}{245, 245, 245}   
\definecolor{promptborder}{RGB}{200, 200, 200} 
\tcbset{
    promptstyle/.style={
        enhanced,
        frame hidden,
        boxrule=0pt,
        left=8pt,
        right=8pt,
        top=6pt,
        bottom=6pt,
        arc=3pt,
        colback=promptbg,
        colframe=promptborder,
        fontupper=\fontsize{0.7em}{0.7em}\selectfont\ttfamily\leftskip,
        before upper={\parindent}, 
        overlay unbroken and first={
            \draw[promptborder, line width=1pt] 
                (frame.south west) -- (frame.north west);
            \draw[promptborder, line width=1pt] 
                ([xshift=-2pt]frame.north east) -- (frame.south east);
        },
        overlay middle and last={
            \draw[promptborder, line width=1pt] 
                (frame.south west) -- (frame.north west);
            \draw[promptborder, line width=1pt]
                ([xshift=-2pt]frame.north east) -- (frame.south east);
        }
    }
}

\usepackage[textsize=tiny]{todonotes}

\definecolor{nred}{RGB}{196, 38, 11}
\definecolor{ngreen}{RGB}{18, 141, 21}
\definecolor{nblue}{RGB}{41, 52, 190}

\newcommand{\ignore}[1]{}

\colmfinalcopy

\newcommand{\method}[0]{\textsc{RLVMR}}

\title{\vspace{-10pt}{\em Beyond Blind Success}: Reinforcement Learning with Verifiable {\color{nblue} Meta-Reasoning} Rewards for Long-Horizon Agents\vspace{-15pt}}

\title{{\em RLVMR}: Reinforcement Learning with Verifiable {\color{nblue} Meta-Reasoning} Rewards for Robust Long-Horizon Agents}

\author[ ]{Zijing Zhang}
\author[ ]{Ziyang Chen$^*$}
\author[ ]{Mingxiao Li}
\author[ ]{Zhaopeng Tu\thanks{Correspondence to: Zhaopeng Tu \textless zptu@tencent.com\textgreater~and Ziyang Chen\textless willzychen@tencent.com\textgreater.}}
\author[ ]{Xiaolong Li}

\affil[ ]{Hunyuan AI Digital Human, Tencent \protect\\[2pt] 
\url{https://github.com/Tencent/DigitalHuman/tree/main/RLVMR}}

\begin{document}

\maketitle

\begin{figure}[ht]
  \centering
  \subfloat[Model Performance on ALFWorld]{\includegraphics[width=0.48\linewidth]{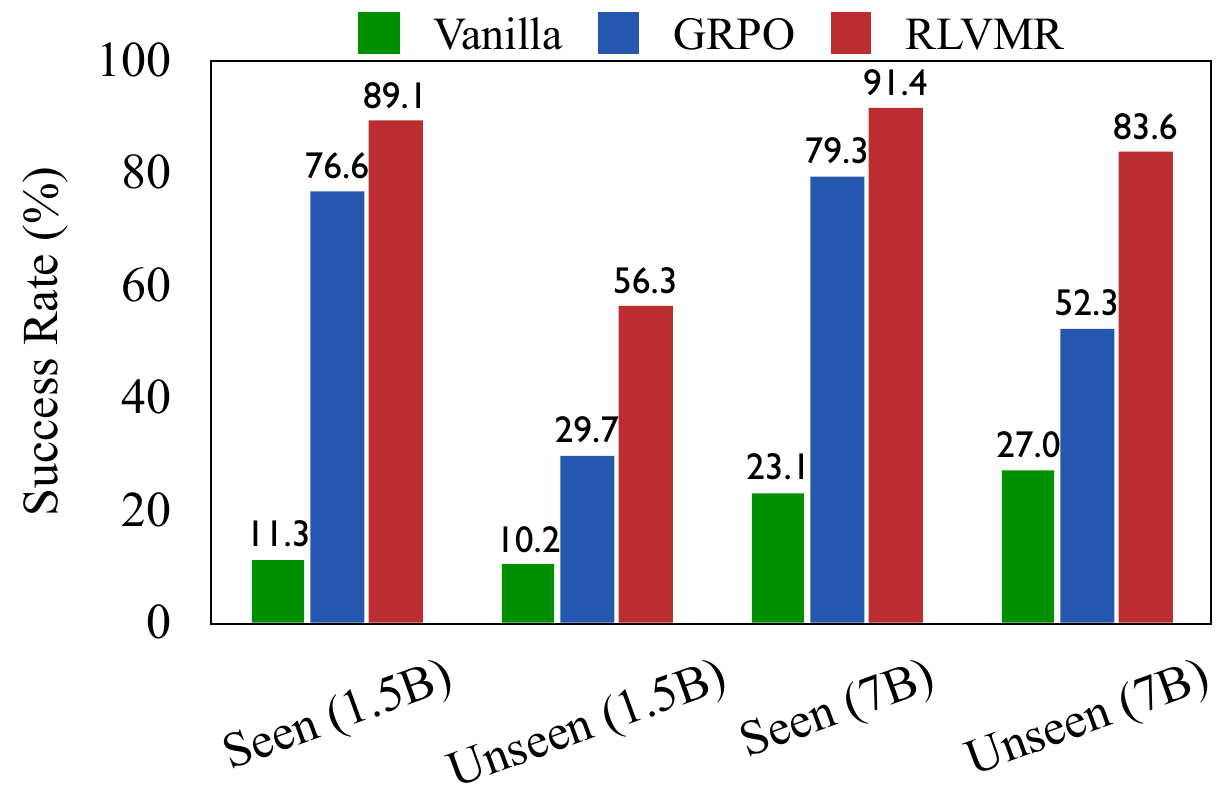}} \hfill
  \subfloat[Repetitive Action Rate]{\includegraphics[width=0.48\linewidth]{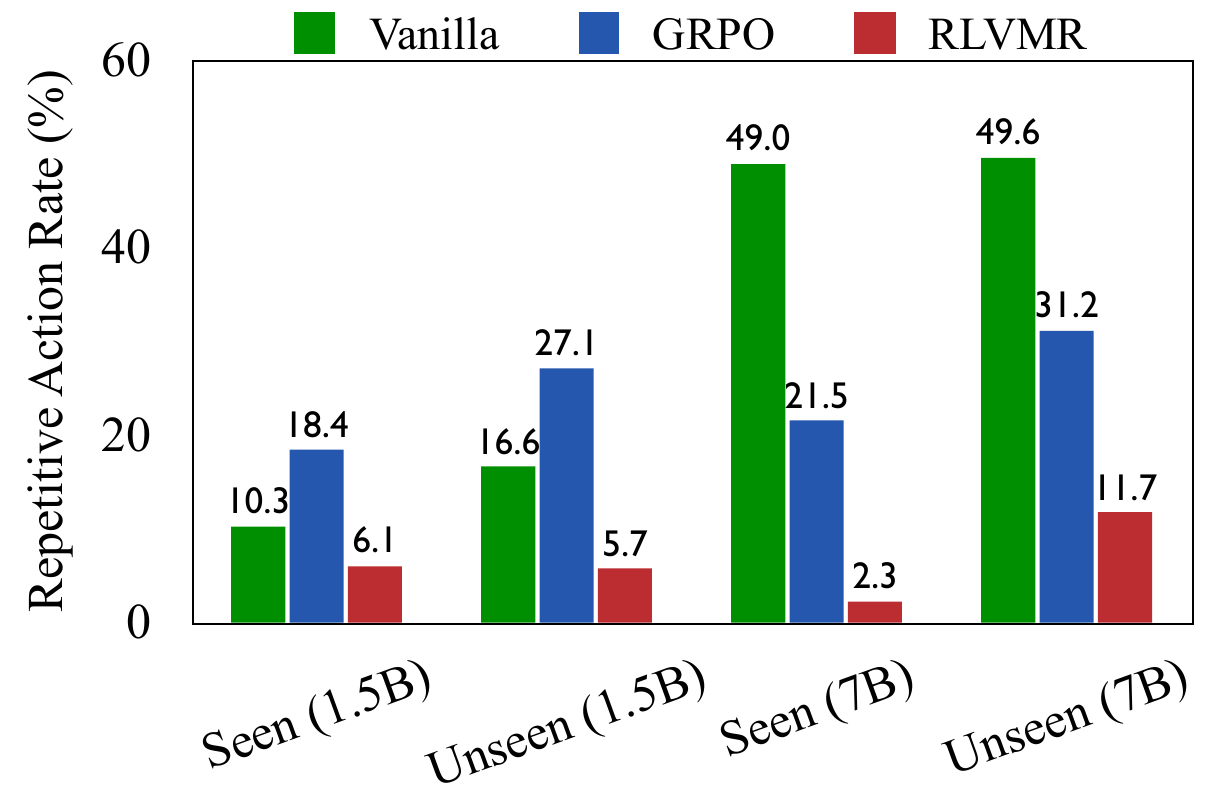}}\\
  \caption{Reinforcement learning with outcome-only rewards (e.g., GRPO) improves performance over vanilla models but fosters inefficient exploration, characterized by high rates of repetitive actions that hinder generalization to unseen tasks. In contrast, our proposed \method{} significantly improves success rates and generalization by directly mitigating this inefficient exploration.}
  \label{fig:front_page}
\end{figure}

\begin{abstract}
The development of autonomous agents for complex, long-horizon tasks is a central goal in AI. However, dominant training paradigms face a critical limitation: reinforcement learning (RL) methods that optimize solely for final task success often reinforce flawed or inefficient reasoning paths, a problem we term {\bf inefficient exploration}. This leads to agents that are brittle and fail to generalize, as they learn to find solutions without learning {\em how} to reason coherently. To address this, we introduce {\bf \method{}}, a novel framework that integrates dense, process-level supervision into end-to-end RL by rewarding verifiable, meta-reasoning behaviors. \method{} equips an agent to explicitly tag its cognitive steps—such as planning, exploration, and reflection—and provides programmatic, rule-based rewards for actions that contribute to effective problem-solving. These process-centric rewards are combined with the final outcome signal and optimized using a critic-free policy
gradient method. On the challenging ALFWorld and ScienceWorld benchmarks, \method{} achieves new state-of-the-art results, with our 7B model reaching an 83.6\% success rate on the most difficult unseen task split. Our analysis confirms these gains stem from improved reasoning quality, including significant reductions in redundant actions and enhanced error recovery, leading to more robust, efficient, and interpretable agents.
\end{abstract}


\clearpage

\section{Introduction}

\begin{figure}[ht]
  \centering
  \subfloat[Standard RLVR (GRPO)]{\includegraphics[width=0.45\linewidth]{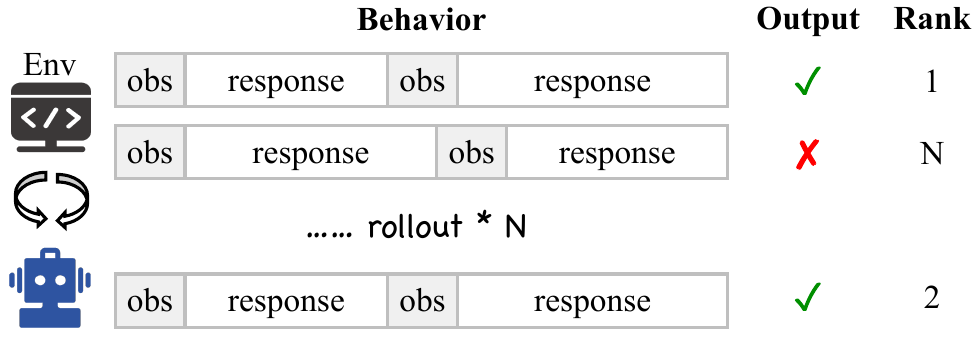}} \hspace{0.05\linewidth}
  \subfloat[\method{} (Ours)]{\includegraphics[width=0.45\linewidth]{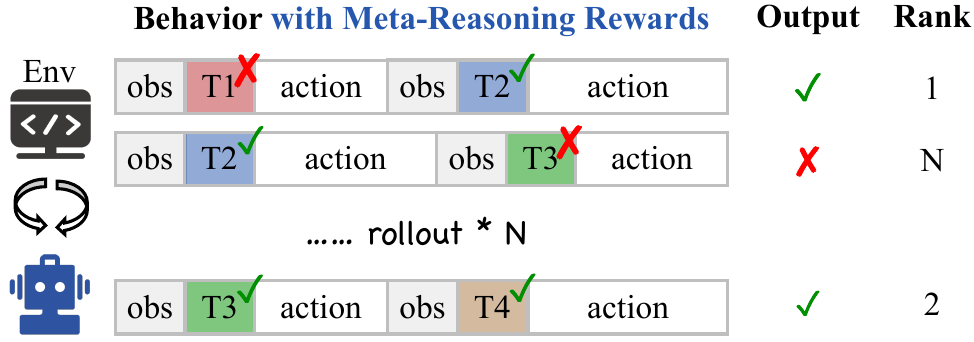}}\\
  \caption{Comparison of LLM agent RL training paradigms: (a) Standard RL with outcome-only rewards (e.g., GRPO) inadvertently reinforces trajectories with inefficient or illogical intermediate reasoning steps. (b) Our \method{} approach provides dense, verifiable rewards for beneficial meta-reasoning behaviors (e.g., T1-T4), directly shaping a more robust and coherent reasoning process.}
  \label{fig:intro}
\end{figure}

The quest to build autonomous agents capable of solving complex, long-horizon tasks has gained significant momentum with the rise of Large Language Models (LLMs)~\citep{zeng2024agenttuning, wang2022scienceworld, bai2024digirl}. However, dominant training paradigms face a fundamental trade-off. On one hand, Supervised Fine-Tuning (SFT) on expert trajectories can teach agents efficient behaviors, but these policies are often brittle and fail to generalize to novel situations~\citep{chu2025sftrl}. On the other hand, Reinforcement Learning (RL) from environmental feedback encourages exploration and can lead to better generalization, but it typically optimizes for a single, sparse reward signal: final task success.

This reliance on outcome-only rewards raises a critical, yet underexplored question: {\bf Are agents learning to reason coherently, or are they just finding brittle shortcuts to success?} Our work investigates a pervasive issue we term {\bf inefficient exploration}, where agents are rewarded for successful outcomes even when their path to success is built on flawed, illogical, or redundant reasoning. As illustrated in Figure~\ref{fig:front_page}, this leads to agents that exhibit high rates of repetitive actions and struggle to adapt to unseen tasks, because their underlying problem-solving process is unsound. Standard RL inadvertently reinforces any successful trajectory, failing to distinguish between robust and flawed reasoning processes. This deficiency undermines agent reliability, interpretability, and generalization, especially as tasks grow in complexity.

We argue that to build truly robust and generalizable agents, we must move beyond rewarding only the final outcome and begin to supervise the reasoning {\em process} itself. Drawing inspiration from metacognitive theory~\citep{martinez2006metacognition}, which posits that effective problem-solving depends on ``thinking about thinking'', we propose to directly reward beneficial cognitive behaviors. Our key insight is that high-level skills like planning, monitoring progress, exploring alternatives, and reflecting on errors can be operationalized as distinct, verifiable steps within an agent's reasoning process.

To this end, we introduce {\bf Reinforcement Learning with Verifiable Meta-Reasoning Rewards (\method{})}, a novel framework that integrates dense, process-level supervision into end-to-end RL. 
As illustrated in Figure~\ref{fig:intro}, \method{} contrasts with standard RL by rewarding not only the final outcome but also the intermediate reasoning steps. 
Our framework defines a set of core meta-reasoning behaviors — {\em planning}, {\em exploration}, and {\em reflection/monitoring} — and enables the agent to articulate its cognitive state through special tags. During online interaction, we use lightweight, programmatic rules to grant verifiable rewards for these behaviors. For example, an `exploration' tag is rewarded when the agent discovers a new state, while a `reflection' tag is rewarded when it leads to the correction of a prior mistake. These process-centric rewards are combined with the global outcome reward and optimized using a policy gradient method. After a brief ``cold-start'' supervised fine-tuning (SFT) phase on only 200 trajectories to learn the tag syntax, the agent is trained entirely through environmental interaction.

We demonstrate the effectiveness of \method{} on two challenging long-horizon benchmarks, ALFWorld and ScienceWorld. Our experiments show that \method{} achieves new state-of-the-art results across all settings. Notably, on the hardest unseen task split (L2), our 7B model achieves an 83.6\% success rate, and surpasses the performance of the much larger models. In-depth analysis reveals that these gains are driven by a tangible improvement in reasoning quality: \method{}-trained agents exhibit significant reductions in repetitive and invalid actions. This confirms that by rewarding the {\em process} of good reasoning, we create agents that are not only more successful but also more robust, efficient, and generalizable.

In summary, our contributions are as follows:
\begin{enumerate}[leftmargin=12pt]
    \item We identify and formulate the {\bf inefficient exploration} problem in long-horizon agents, showing how optimizing for final outcomes alone reinforces flawed reasoning and leads to brittle policies that fail to generalize.
    \item We propose {\bf \method{}}, a novel RL framework that provides dense, process-level supervision by rewarding verifiable meta-reasoning behaviors (e.g., planning, exploration, reflection) using lightweight, programmatic rules.
    \item We achieve {\bf state-of-the-art performance} on the challenging ALFWorld and ScienceWorld benchmarks, with significant improvements in generalization to unseen tasks. 
    \item We provide {\bf in-depth analysis} confirming that \method{}'s gains stem directly from improved reasoning quality, evidenced by measurable reductions in repetitive actions and enhanced error recovery, thereby improving both agent robustness and efficiency.
\end{enumerate}

\section{Inefficient Exploration in Long-Horizon Agents}

This section investigates the phenomenon of ``inefficient exploration'' in agents designed for long-horizon tasks. We analyze its detrimental effects on performance, which manifest as {\bf brittle efficiency} on previously seen tasks and {\bf poor generalization} to unseen ones.

\subsection{Experimental Setup}

\paragraph{Benchmarks}
To evaluate foundational capabilities and generalization, we conduct experiments on the widely-used and challenging {\bf ALFWorld} benchmark~\citep{shridhar2020alfworld}, which comprises embodied household tasks. To systematically measure generalization, we define three evaluation splits based on the original benchmark:
\begin{itemize}[leftmargin=12pt]
    \item \textbf{L0} (\emph{seen-L0}): {\color{ngreen} seen task variants and seen task categories}; 
    \item \textbf{L1} (\emph{unseen-L1}): {\color{nred} unseen held-out task variants} but {\color{ngreen} seen task categories};
    \item \textbf{L2} (\emph{unseen-L2}): {\color{nred} unseen held-out task variants and unseen task categories}.
\end{itemize}
L0 and L1 follow the official benchmark splits. For L2, we further partition ALFWorld by task category, holding out entire categories from training for exclusive use in evaluation.

\paragraph{Training Paradigms}
We experiment with Qwen2.5-1.5B-Instruct and Qwen2.5-7B-Instruct models using the \textbf{ReAct}~\citep{yao2023react} framework, which alternates between reasoning and acting steps. We evaluate two dominant training paradigms:
\begin{itemize}[leftmargin=12pt]
    \item \textbf{SFT}~\citep{yang2023gpt4tools, tang2023toolalpaca, xi2024agentgym}: A widely adopted paradigm that applies supervised fine-tuning on high-quality expert trajectories.
    \item \textbf{GRPO}~\citep{feng2025GRPO, wang2025ragen}: An end-to-end RL method that optimizes the policy by comparing the final rewards of multiple trajectories sampled from the same initial state.
\end{itemize}

\paragraph{Evaluation Metrics}
We assess performance using the following metrics:
\begin{itemize}[leftmargin=12pt]
    \item {\bf Success Rate (\%, $\uparrow$):} The percentage of tasks successfully completed by the agent on each evaluation split.
    \item \textbf{Invalid Action Rate (\%, $\downarrow$):} The proportion of generated actions that are invalid in the current state, reflecting basic comprehension and error frequency.
    \item {\bf Repetitive Action Rate (\%, $\downarrow$):} The percentage of steps where the agent executes a {\bf meaningless repeated action}, as defined in prior work~\citep{yuan2025agentr, fu2025agentrefine, feng2025gigpo}. This metric quantifies inefficient exploration, indicating that the agent's policy may be overfitting to familiar action sequences rather than being guided by robust reasoning.
\end{itemize}

\subsection{Observing the Inefficient Exploration Problem}

Before presenting aggregate statistics, we first examine individual trajectories to understand how inefficient exploration manifests at the step level. Such microscopic inspection reveals the cognitive patterns that lead to brittle or wasteful behaviors. We observe that, although various methods (e.g., GRPO) can improve an agent's success rate on specific tasks, such improvement is often due to reinforcement of state-action mappings associated with correct reasoning. However, the agent's self-reflection and understanding of its own reasoning process are frequently overlooked.

The following is a trajectory segment of a vanilla GRPO-trained agent performing a novel task—-- \textit{put two keychains in the safe} --- in ALFWorld (corresponding to our L2 split). By step 7, the agent has already arrived at dresser 1. However, we observe that in the subsequent steps, the agent falls into a sequence of inefficient decisions: its next intention is to find the second keychain, yet it persistently tries to go to dresser 1 for several steps, disregarding the fact that it is already there. This indicates that its policy mainly reflects the action distribution present in the training data, rather than allowing the reasoning process itself to truly regulate decision-making. Although the agent can form relatively effective action strategies for completing tasks, its capacity for critically evaluating its own behavior and understanding the underlying task requirements remains limited. This suggests that the agent has not truly acquired the reasoning patterns necessary for robust task-solving.

\begin{tcolorbox}[title = {A trajectory segment of an agent trained with GRPO}, breakable]
\textbf{Step 7}

\textbf{Observation:} You arrive at dresser 1. On the dresser 1, you see a box 1, a creditcard 1, a keychain 2, a keychain 1, a laptop 1, a statue 1, and a tissuebox 2.\\
\textbf{Reasoning:} The current situation is that I am at dresser 1, which has multiple items including a keychain 1, a keychain 2, a creditcard 1, etc. To complete the task of putting two keychains in the safe, I need to select one of these keychains from the dresser.\\
\textbf{Action:} take keychain 1 from dresser 1\\

\textbf{Step 8}

\textbf{Observation:} You pick up the keychain 1 from the dresser 1.\\
\textbf{Reasoning:} The current situation is that I am at dresser 1 with the keychain 1 in my possession. To put two keychains in the safe, I need to find another keychain and then move both keychains to the safe.\\
\textbf{Action:} go to dresser 1\\

\textbf{Step 9}

\textbf{Observation:} Nothing happens.\\
\textbf{Reasoning:} The player has been in the room for several steps but has not yet interacted with the safe or taken any keychains. The safe is closed initially, which might need interaction to open it.\\
\textbf{Action:} go to dresser 1\\

\textbf{Step 10}

\textbf{Observation:} Nothing happens.\\
\textbf{Reasoning:} The previous steps were not successful in opening the safe or taking the keychains. Let's try another approach.\\
\textbf{Action:} go to dresser 1\\

\textbf{Step 11}

\textbf{Observation:} Nothing happens.\\
\textbf{Reasoning:} The user has taken all the steps necessary to obtain both keychains but has yet to actually put them into the safe. The last known location of the keychains was on the dresser before the attempt to open the safe failed (in Observation 8). To complete the task, the user needs to move one or both keychains to the safe before attempting to open it again.\\
\textbf{Action:} move keychain 1 to safe 1\\

\textbf{Observation:} Nothing happens.\\
\end{tcolorbox}

This qualitative example highlights how an ostensibly ``successful'' policy can still waste a large number of steps on redundant or illogical moves.  We next quantify how pervasive this phenomenon is across models and training paradigms.

\subsection{The Problem of Inefficient Exploration}

\begin{figure}[t]
  \centering
  \subfloat[Success Rate (1.5B)]{\includegraphics[width=0.3\linewidth]{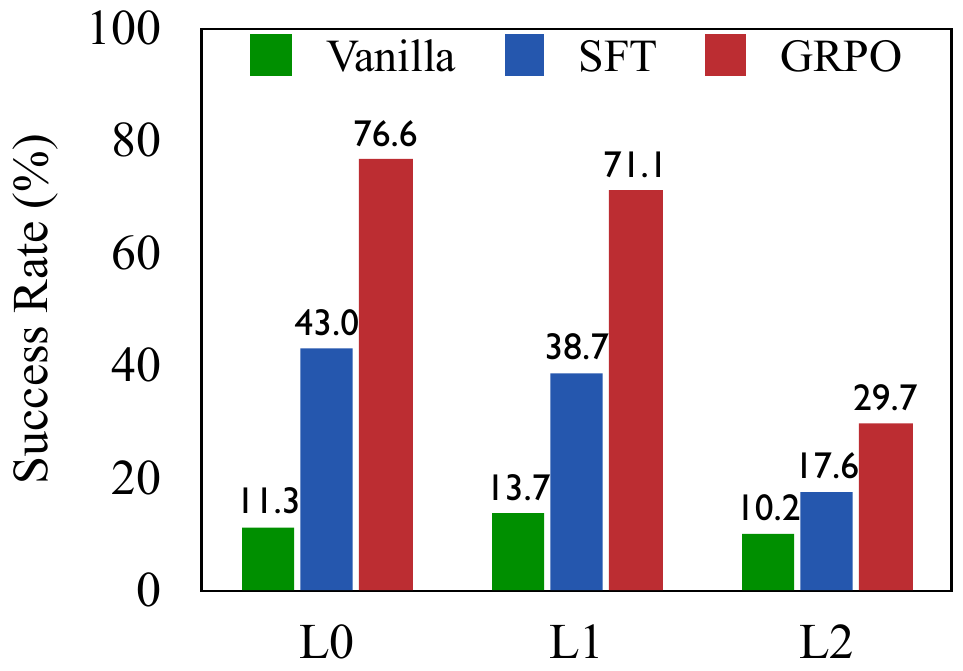}} \hfill
  \subfloat[Invalid Actions (1.5B)]{\includegraphics[width=0.3\linewidth]{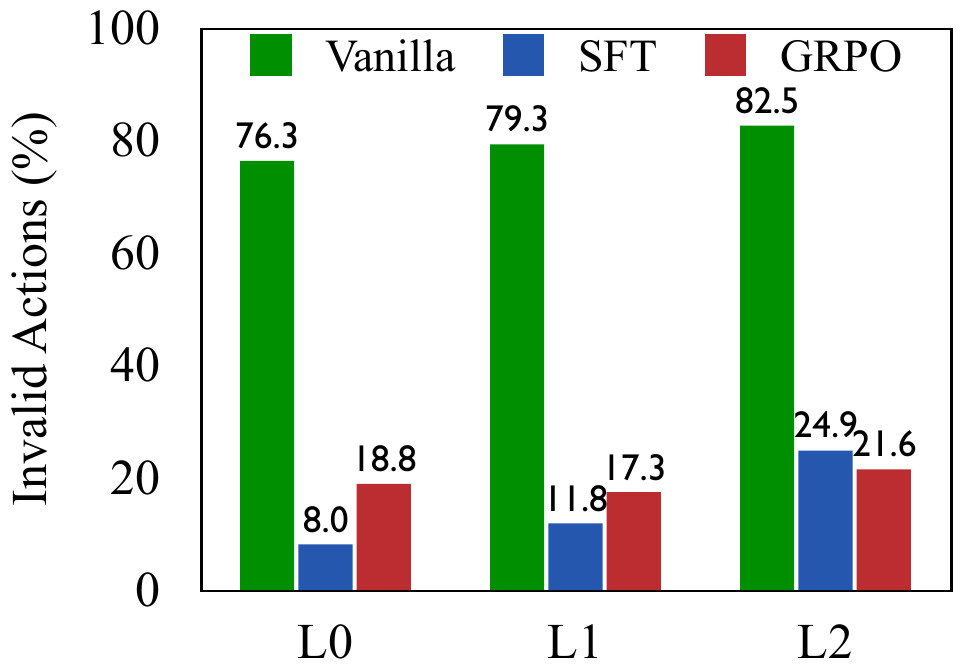}} \hfill
  \subfloat[Repetitive Actions (1.5B)]{\includegraphics[width=0.3\linewidth]{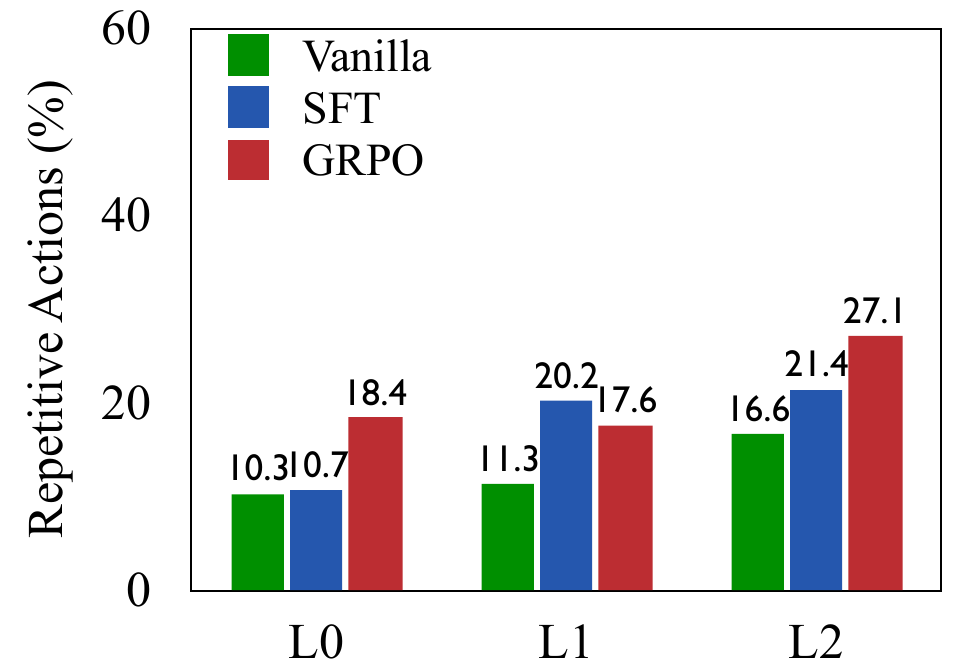}}\\\vspace{5pt}
  \subfloat[Success Rate (7B)]{\includegraphics[width=0.3\linewidth]{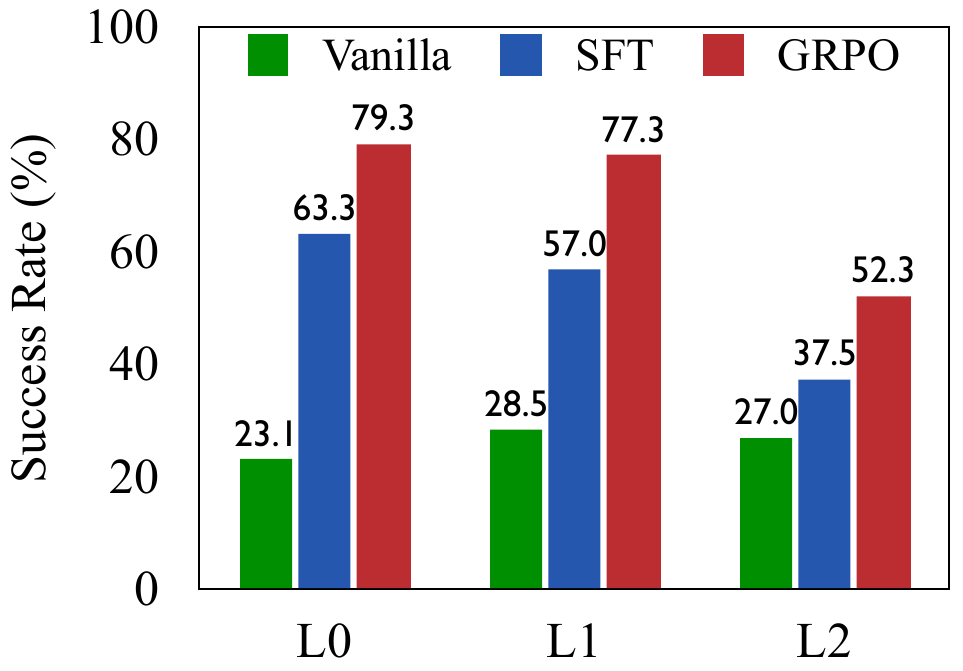}} \hfill
  \subfloat[Invlaid Actions (7B)]{\includegraphics[width=0.3\linewidth]{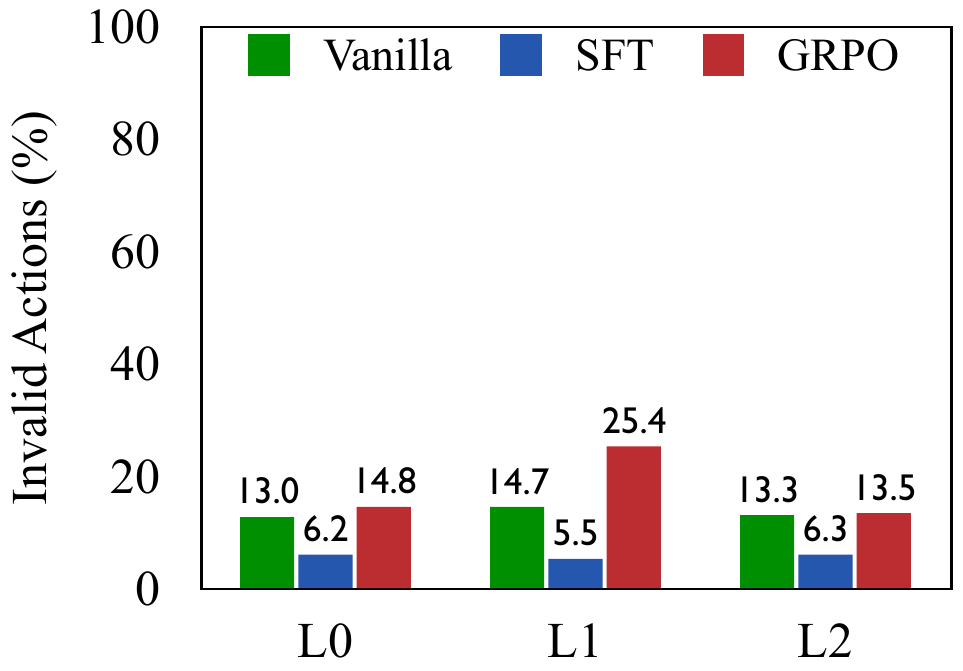}} \hfill
  \subfloat[Repetitive Actions (7B)]{\includegraphics[width=0.3\linewidth]{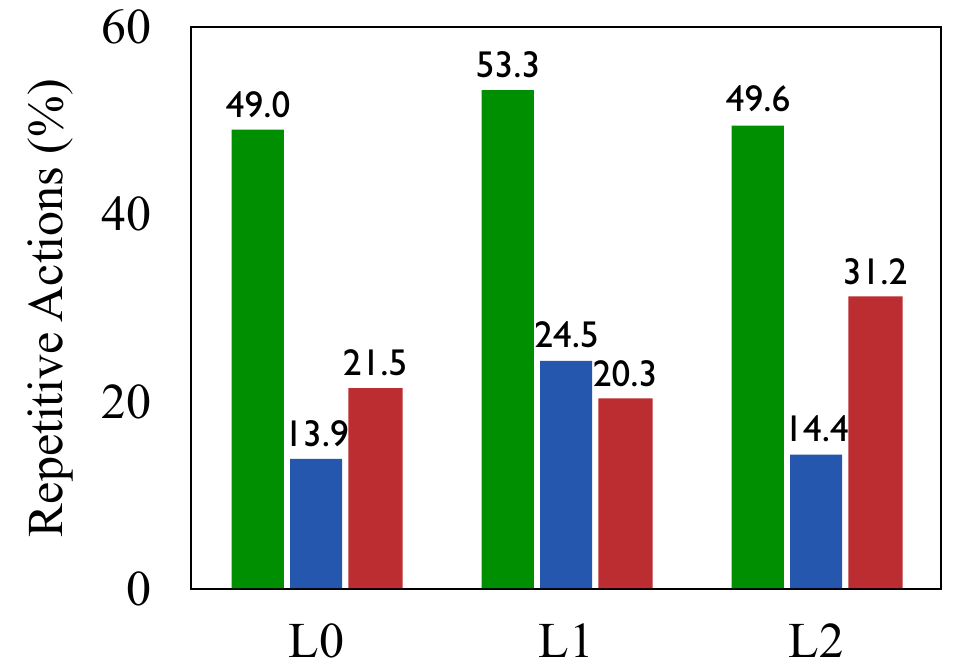}}
\caption{Performance of SFT and GRPO on ALFWorld. While SFT excels on seen tasks (L0) but fails to generalize, GRPO achieves better generalization at the cost of significant inefficiency (high action counts and redundancy). This highlights a fundamental trade-off between brittle efficiency and inefficient generalization.}
  \label{fig:inefficient-exploration}
\end{figure}

We now present large‐scale empirical results that corroborate the anecdotal evidence above. Figure~\ref{fig:inefficient-exploration} compares SFT and GRPO across success, invalid‐action, and repetitive‐action metrics.

\paragraph{SFT creates efficient but brittle policies that fail to generalize.}
As seen, SFT significantly boosts performance on seen tasks (L0) compared to the ReAct baseline. For instance, the 7B model's success rate jumps from 23.1\% to 63.3\%. This approach also yields highly efficient policies with low invalid action rates (e.g., 6.2\% on L0 for the 7B model). However, this efficiency is brittle. On the most challenging unseen split (L2), the 7B model's success rate plummets to 37.5\%. Furthermore, its repetitive action rate nearly doubles from 13.9\% on L0 to 24.5\% on L1, revealing a critical flaw: when faced with novel situations not covered by expert data, the agent falls back on non-productive loops. This demonstrates that SFT teaches agents to mimic actions without instilling a robust, generalizable reasoning process.

\paragraph{RL with outcome-only rewards (GRPO) improves generalization but fosters inefficient and flawed reasoning.}
In contrast, GRPO achieves substantially better generalization. The 7B GRPO model attains success rates of 77.3\% on L1 and 52.3\% on L2, significantly outperforming SFT. This success, however, validates our core hypothesis about the {\bf inefficient exploration problem}. The agent's performance is undermined by severe inefficiency, as evidenced by high invalid and repetitive action rates across all splits. For example, the 7B model's repetitive action rate on the most difficult L2 tasks is a staggering 31.2\%. By optimizing solely for final task success, GRPO reinforces any path that leads to a positive outcome, even those built on illogical steps, redundant actions, and inefficient exploration.

\paragraph{Scaling the model size improves baseline capabilities but does not fix the underlying reasoning deficiencies.}
While scaling from a 1.5B to a 7B model improves overall success rates for both SFT and GRPO, it does not resolve the fundamental issue. The 7B GRPO model, despite its higher success rate on L2 (52.3\% vs. 29.7\% for 1.5B), also exhibits a higher repetitive action rate (31.2\% vs. 27.1\%). This indicates that a larger model's enhanced capabilities can sometimes be misdirected to more effectively exploit flawed strategies, rather than to reason more coherently. This finding underscores that the limitations are rooted in the training objective itself, not merely in the model's capacity. Simply increasing model size is not a panacea for the inefficient exploration problem.

\paragraph{Current training paradigms force a trade-off between brittle efficiency and inefficient generalization.}
Our analysis reveals a fundamental dilemma in agent training. SFT produces policies that are efficient on familiar tasks but brittle and unable to generalize, as they lack robust problem-solving skills. Conversely, GRPO fosters better generalization through exploration but at the cost of reinforcing inefficient and logically flawed reasoning paths. Neither paradigm effectively teaches the agent {\em how} to reason well. This establishes a clear need for a new framework that moves beyond sparse, outcome-only signals to provide direct, {\bf process-level supervision}. By rewarding coherent and efficient reasoning steps, we can guide agents to not only find solutions but to do so robustly and intelligently, which is the precise goal of our RLVMR framework.

\section{Methodology}

\begin{figure*}[t]
  \centering
    \includegraphics[width=\linewidth]{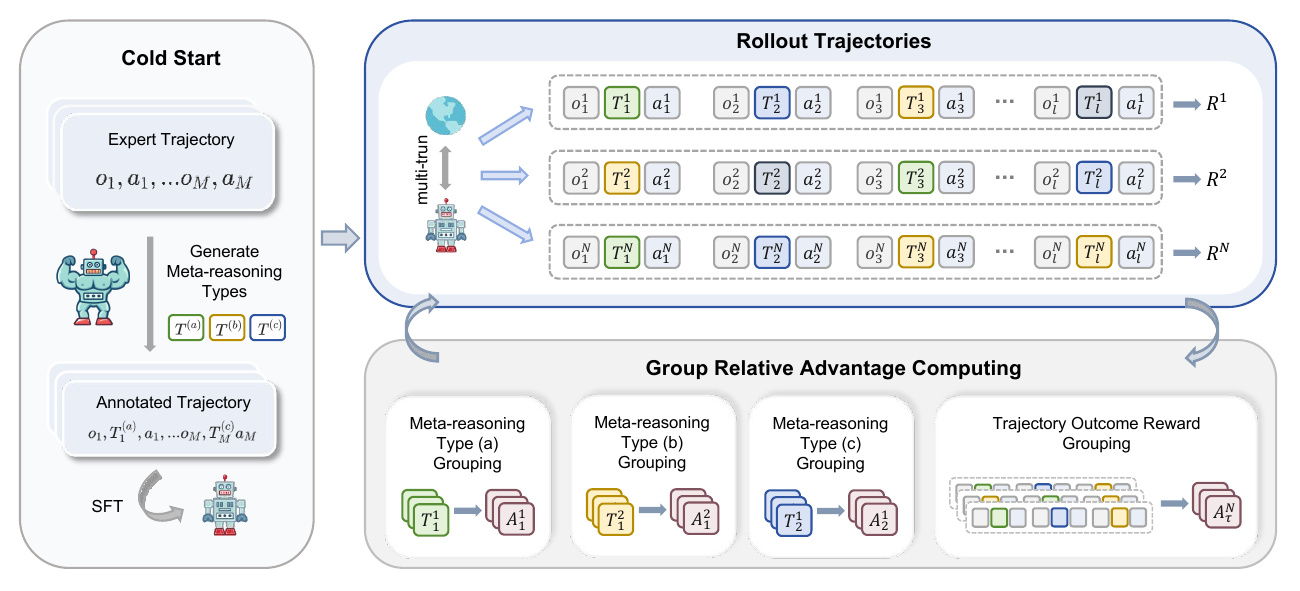}
    \caption{A schematic diagram of the RLVMR framework, which consists of two training phases: cold start and reinforcement learning. Our method provides rule-verifiable feedback signals based on the final outcome and the relative advantages of different types of meta-reasoning behaviors.
    }
    \label{fig:framework}
\end{figure*}

Our methodology equips LLM agents with an explicit meta-reasoning framework to mitigate inefficient exploration in complex tasks. We begin by formalizing the agent-environment interaction as a Markov Decision Process (\S~\ref{sec:mdp}). We then introduce a novel meta-reasoning framework that extends existing agent architectures by operationalizing principles from cognitive science (\S~\ref{sec:operations}). As illustrated in Figure~\ref{fig:framework}, the agent is trained in two phases: an initial SFT stage to bootstrap the agent's meta-reasoning capabilities (\S~\ref{sec:cold_start}), followed by a reinforcement learning phase that uses a custom policy optimization algorithm to refine these skills based on task outcomes and process-centric rewards (\S~\ref{sec:rlvmr}).

\subsection{Task Formulation as a Markov Decision Process}
\label{sec:mdp}

We formalize the interaction between an agent and its environment in long-horizon tasks as a Markov Decision Process (MDP). An MDP is defined by a tuple $(S, A, O, F, R)$, where $S$ is the set of environment states, $A$ is the action space, $O$ is the observation space, $F: S \times A \rightarrow S$ is the state transition function, and $R: S \times A \rightarrow \mathbb{R}$ is the reward function. In our setting, which is tailored for LLM agents, the state, action, and observation spaces ($S, A, O$) are all represented as natural language sequences over a finite token vocabulary.

At each timestep $t$, the agent's policy $\pi_\theta$ generates a thought process $th_t$ and an action $a_t$ based on the current state $s_t$: $(th_t, a_t) \sim \pi_\theta(\cdot \mid s_t)$. The agent's interaction with the environment produces a trajectory $\tau = \{(o_1, th_1, a_1), (o_2, th_2, a_2), \ldots, (o_n, th_n, a_n)\}$. In many long-horizon tasks, reward signals are sparse, typically provided only as a final outcome reward $R(\tau)$ at the end of an episode. This sparsity poses significant challenges for credit assignment. The agent's objective is to learn an optimal policy $\pi_\theta$ that maximizes the expected cumulative reward:
\begin{equation}
    \max_\theta\ \mathbb{E}_{\tau \sim \pi_\theta} \left[ R(\tau) \right].
\end{equation}

\subsection{Operationalizing Meta-Reasoning in LLM Agents}
\label{sec:operations}

Our approach is grounded in metacognitive theory~\citep{martinez2006metacognition, lai2011metacognition}, which emphasizes ``thinking about thinking''. Metacognition comprises two key components: \textit{metacognitive knowledge} (an agent's self-awareness of its own reasoning strategies) and \textit{metacognitive regulation} (the active control of these processes, including planning, monitoring, and adaptive revision). This theoretical lens suggests that for LLM agents to solve complex tasks, they require not just domain knowledge but also the capacity for dynamic planning, self-monitoring, and creative exploration.

To operationalize these principles, we extend the ReAct framework. While ReAct interleaves reasoning and actions (e.g., ``Think: ..., Act: ...''), it treats reasoning as a monolithic process. We refine this by introducing a structured set of meta-reasoning tags to explicitly represent distinct cognitive functions. This decouples reasoning from actions and enables fine-grained analysis and supervision. Specifically, we define four meta-reasoning tags, each enclosed in XML-style tags (e.g., \texttt{<planning>}), while all actions are contained within the \texttt{<action>} tag.

\begin{itemize}
    \item \textbf{Planning (\texttt{<planning>}):} Decomposes the task into high-level steps to formulate an overall strategy. Used at the start of a task or when replanning is needed.
    \item \textbf{Exploration (\texttt{<explore>}):} Generates hypotheses or options to navigate uncertainty or bottlenecks, encouraging creative problem-solving.
    \item \textbf{Reflection (\texttt{<reflection>}):} Reviews history to analyze errors and formulate corrective actions. Typically triggered after unsuccessful attempts.
    \item \textbf{Monitoring (\texttt{<monitor>}):} Tracks task progress against the overall plan, ensuring actions remain aligned with subgoals. Applied during routine execution.
\end{itemize}

\subsection{Cold Start: Initial Meta-Reasoning Acquisition via SFT}
\label{sec:cold_start}

To equip the base LLM with the foundational ability to generate structured meta-reasoning, we begin with a supervised fine-tuning phase. This step is crucial, as reasoning patterns learned during subsequent reinforcement learning are heavily influenced by the base model's capabilities. The SFT data is constructed as follows:
\begin{enumerate}
    \item We collect a dataset of successful task trajectories containing only observation-action pairs.
    \item We employ a more powerful teacher model (e.g., GPT-4) to annotate these trajectories with our meta-reasoning tags, inferring the most likely cognitive step preceding each action. This process creates synthetic, reasoning-rich expert demonstrations.
    \item The target LLM is fine-tuned on these annotated trajectories, learning to imitate the expert's meta-reasoning and action generation patterns.
\end{enumerate}

\subsection{\method}
\label{sec:rlvmr}

\subsubsection{Meta-Reasoning-Aware Reward Shaping}
\label{sec:reward_shaping}

During reinforcement learning, we guide the agent with a composite reward signal that combines task completion with the quality of the reasoning process. This signal comprises a sparse outcome reward and a dense, process-based meta-reasoning reward.

\textbf{Outcome Reward ($R(\tau)$):} A binary signal awarded at the end of a trajectory: $R(\tau) = r_s$ for task success and $0$ otherwise, where $r_s$ is a positive constant.

\textbf{Meta-Reasoning Reward ($r_t^{\mathrm{MR}}$):} A dense reward assigned at each step $t$ to incentivize locally beneficial behaviors.
\begin{itemize}
    \item \textbf{Planning Reward ($r_{\mathrm{planning}}$):} Awarded for a \texttt{<planning>} step if the trajectory ultimately succeeds.
    \item \textbf{Exploration Reward ($r_{\mathrm{explore}}$):} Awarded if the current action targets a new object or location, discouraging redundancy.
    \item \textbf{Reflection Reward ($r_{\mathrm{reflection}}$):} Awarded if a \texttt{<reflection>} step is followed by a corrective action after a sequence of failures.
\end{itemize}

\textbf{Format Reward ($r_t^{\mathrm{format}}$):} A penalty, $-\lambda_{\mathrm{format}}$, is applied if the model's output at step $t$ does not conform to the expected \texttt{<tag>...<action>...</action>} structure.

The total step-level reward is the sum of the process-based rewards: $r_t = r_t^{\mathrm{MR}} + r_t^{\mathrm{format}}$.

\subsubsection{Group Relative Policy Optimization with Meta-Reasoning (GRPO-MR)}
\label{sec:GRPO_MR}

To effectively leverage our composite reward signal, we introduce Meta-Reasoning Group Policy Optimization (GRPO-MR), an algorithm adapted from PPO. GRPO-MR computes a step-level advantage by combining global trajectory performance with local, context-aware reasoning quality.

\textbf{Trajectory-level Relative Advantage:} For a batch of $K$ trajectories, we first calculate a normalized trajectory-level advantage to capture overall performance:
\begin{equation}
    A^{\mathrm{traj}}_k = \frac{R(\tau_k) - \mu_R}{\sigma_R},
\end{equation}
where $\mu_R$ and $\sigma_R$ are the mean and standard deviation of outcome rewards across the batch.

\textbf{Meta-reasoning Level Relative Advantage:} The core of GRPO-MR is the computation of a context-aware advantage. We group all steps within a batch that share the same meta-reasoning tag (e.g., all \texttt{<explore>} steps) and normalize their rewards \textit{within} that group:
\begin{equation}
    A^{\mathrm{MR}}_{t,\mathrm{tag}} = \frac{r^{\mathrm{MR}}_{t,\mathrm{tag}} - \mu_{\mathrm{tag}}}{\sigma_{\mathrm{tag}}},
\end{equation}
where $\mu_{\mathrm{tag}}$ and $\sigma_{\mathrm{tag}}$ are the mean and standard deviation of meta-reasoning rewards for all steps with that specific `tag`.

The final step-level advantage $A_t$ is a weighted combination of these two signals:
\begin{equation}
    A_t = \alpha \cdot A^{\text{traj}}_k + (1 - \alpha) \cdot A^{\text{MR}}_{t, \text{tag}},
\end{equation}
where $\alpha \in [0, 1]$ is a hyperparameter balancing the influence of the global outcome and local reasoning quality.

Finally, we optimize the policy $\pi_\theta$ using a clipped surrogate objective with KL divergence regularization:
\begin{equation}
\mathcal{L}_{\text{final}} = \mathbb{E}_t \left[ \min \left( r_t(\theta) A_t, \text{clip}(r_t(\theta), 1-\epsilon, 1+\epsilon)A_t \right) \right] - \lambda_{\text{KL}} D_{\text{KL}}(\pi_\theta \Vert \pi_{\text{ref}}),
\end{equation}
where $r_t(\theta)$ is the importance sampling ratio, $\epsilon$ is the clipping hyperparameter, and $\lambda_{\text{KL}}$ controls the KL penalty against a reference policy $\pi_{\text{ref}}$.

\section{Experiment}

\begin{table*}[t]
\centering
\setlength{\tabcolsep}{8pt}
\caption{Performance comparison on the ALFWorld and ScienceWorld benchmarks. We report the success rate (\%) on seen (L0: {\color{ngreen} seen task variants and categories}) and unseen (L1: {\color{nred} unseen task variants} but {\color{ngreen} seen task categories}; L2: {\color{nred} unseen task variants and categories}) task variations. Our method, RLVMR, consistently outperforms all baselines across both benchmarks and model sizes.}
\begin{tabular}{llrrrrrr}
\toprule
\multirow{2}{*}{\textbf{Model}}
& \multirow{2}{*}{\textbf{Method}}
& \multicolumn{3}{c}{\textbf{ALFWorld}}
& \multicolumn{3}{c}{\textbf{ScienceWorld}} \\
\cmidrule(lr){3-5} \cmidrule(lr){6-8}
& & \bf L0 & \bf L1 & \bf L2 & \bf L0 & \bf L1 & \bf L2\\
\midrule
AgentGym & SFT+RL & 76.6 & 63.3 & -- & 46.9 & 33.6 & -- \\
\hdashline
GPT-4o   & \multirow{3}{*}{ReAct} & 57.3 & 66.0 & 68.8 & 45.4 & 49.2 & 41.0 \\
DeepSeek-V3 & & 60.2 & 65.9 & 53.9 & 27.3 & 35.2 & 26.5 \\
DeepSeek-R1 & & 68.8 & 70.2 & 67.3 & 22.2 & 31.4 & 29.1  \\
\midrule
\multirow{6}{*}{Qwen-1.5B}
 & ReAct & 11.3 & 13.7 & 10.2 & 1.2 & 0.8 & 0.8 \\
 & +~SFT   & 43.0 & 38.7 & 17.6 & 20.3 & 18.0 & 12.5 \\
 & +~ETO  & 64.1 & 66.4 & 25.8 & 39.1 & 22.7 & 15.6 \\
 & +~GRPO  & 76.6 & 71.1 & 29.7 & 21.1 & 13.7 & 10.9 \\
 & +~GiGPO & 86.7 & 83.2 & 48.0 & 25.8 & 15.2 & 4.7 \\
 & +~RLVMR  & \textbf{89.1} & \textbf{87.9} & \textbf{56.3} & \textbf{46.9} & \textbf{34.4} & \textbf{26.5} \\
\midrule
\multirow{6}{*}{Qwen-7B}
 & ReAct & 23.1 & 28.5 & 27.0 & 7.8 & 11.3 & 6.3 \\
 & +~SFT   & 63.3 & 57.0 & 37.5 & 36.7 & 32.0 & 23.4 \\
 & +~ETO  & 70.3 & 74.2 & 51.6 & 62.5 & 40.6 & 28.1 \\
 & +~GRPO  & 79.3 & 77.3 & 52.3 & 49.1 & 30.1 & 26.6 \\
 & +~GiGPO & 89.5 & 90.2 & 67.2 & 53.4 & 35.2 & 25.8 \\
 & +~RLVMR  & \textbf{91.4} & \textbf{91.8} & \textbf{83.6} & \textbf{67.2} & \textbf{43.0} & \textbf{32.2} \\
\bottomrule
\end{tabular}
\label{tab:main-results}
\end{table*}

\subsection{Main Results}

In this section, we present the core experimental results to evaluate the effectiveness of our proposed \method.
In addition to {\bf ALFWorld}, we also conduct experiments on {\bf ScienceWorld}~\citep{wang2022scienceworld}, which focuses on text-based scientific experimentation.
We evaluate our approach against two more commonly-used training paradigms in addition to SFT and GRPO:
\begin{enumerate}[leftmargin=12pt]
    \item \textbf{ETO}~\citep{song2024ETO}: A RL method that iteratively refines actions using step-level feedback along trajectories.
    \item \textbf{GiGPO}~\citep{feng2025gigpo}: A competitive method that introduces a two-level structure for finer-grained credit assignment.
\end{enumerate}
For broader comparison, we also report the performance of GPT-4o, DeepSeek-V3/R1 and AgentGym~\citep{xi2024agentgym}. AgentGym is trained on Llama-2-Chat-7B, first with behavior cloning on the AgentTraj~\citep{xi2024agentgym} dataset from multiple environments, and then further improved via exploration and self-evolution on a broader instruction set.
For the cold-start phase, we perform supervised fine-tuning on 200 trajectories for 5 epochs. In the RL phase, we set the weighting parameter to $\alpha = 0.5$ and apply a penalty of $-0.1$ to the reward for format violations. The maximum number of steps per episode is fixed at 30 for both benchmarks. Our method is trained for 100 epochs in the RL stage, whereas RL-based baselines are trained for 150 epochs.
Detailed information is provided in Appendix~\ref{app:setup}.

Table~\ref{tab:main-results} presents the results, where we have several findings.

\paragraph{\method{} achieves new SOTA performance across all benchmarks and model sizes.}
Our proposed \method{} framework consistently sets a new standard for performance, outperforming all baseline methods on both ALFWorld and ScienceWorld. With the Qwen-7B model, \method{} achieves success rates of 91.4\% on seen ALFWorld tasks and 67.2\% on seen ScienceWorld tasks, surpassing the next-best method, GiGPO. This consistent superiority highlights the broad applicability and effectiveness of integrating verifiable meta-reasoning rewards into the RL training loop, leading to more capable and successful agents.

\paragraph{Rewarding meta-reasoning significantly enhances generalization to unseen tasks.}
A primary contribution of this work is addressing the inefficient exploration issue to improve generalization. Our results validate this claim, showing that \method{} excels in novel scenarios, especially on the most challenging Unseen-L2 split, which involves entirely new task categories. On ALFWorld's L2 split, our 7B model reaches an impressive 83.6\% success rate, a substantial 16.4 percentage point improvement over the strongest baseline (GiGPO). Similarly, on ScienceWorld's L2 split, \method{} outperforms all other methods. This demonstrates that by learning {\bf how} to reason effectively—rather than just memorizing solutions—our agent develops more robust and transferable problem-solving skills, leading to superior performance on unfamiliar challenges.

\paragraph{\method{} enables smaller models to outperform much larger ones like GPT-4o and DeepSeek-V3/R1.}
The efficiency of our approach is underscored by the performance of our smaller models. The Qwen-1.5B model, when trained with \method{}, achieves a success rate of 87.9\% on the Unseen-L1 split of ALFWorld, decisively outperforming the much larger and more powerful GPT-4o, which scored 66.0\% using a standard ReAct prompting strategy. This result powerfully illustrates that targeted, process-level supervision through our meta-reasoning rewards is a more effective and efficient path to high performance than relying solely on the scale capabilities of massive pre-trained models.

\paragraph{The performance gains are directly attributable to the novel verifiable meta-reasoning rewards.}
To isolate the impact of our core contribution, we compare \method{} against Vanilla-GRPO and GiGPO, which also use trajectory-level optimization but lack process-level rewards. The performance gap is stark. On the ALFWorld L2 split with the 1.5B model, \method{} (56.3\%) achieves nearly double the success rate of Vanilla-GRPO (29.7\%) and significantly surpasses GiGPO (48.1\%). Since \method{} builds upon the same GRPO foundation, this substantial improvement can be directly attributed to the dense, verifiable rewards for beneficial reasoning behaviors. This confirms our central hypothesis: explicitly rewarding the {\bf process} of good reasoning, not just the final {\bf outcome}, is the key driver of \method{}'s superior performance and robustness.

\subsection{Exploration Efficiency}

\begin{figure}[t]
  \centering
  \subfloat[Invalid Actions (1.5B)]{\includegraphics[width=0.4\linewidth]{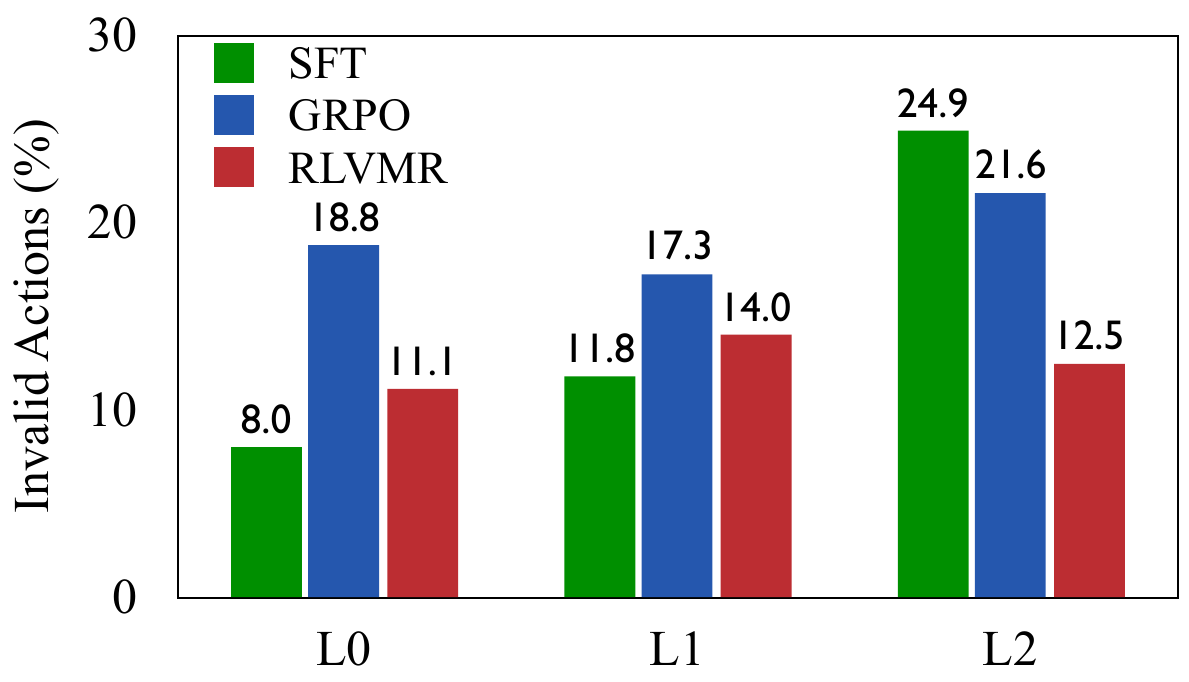}} \hspace{0.1\linewidth}
  \subfloat[Repetitive Actions (1.5B)]{\includegraphics[width=0.4\linewidth]{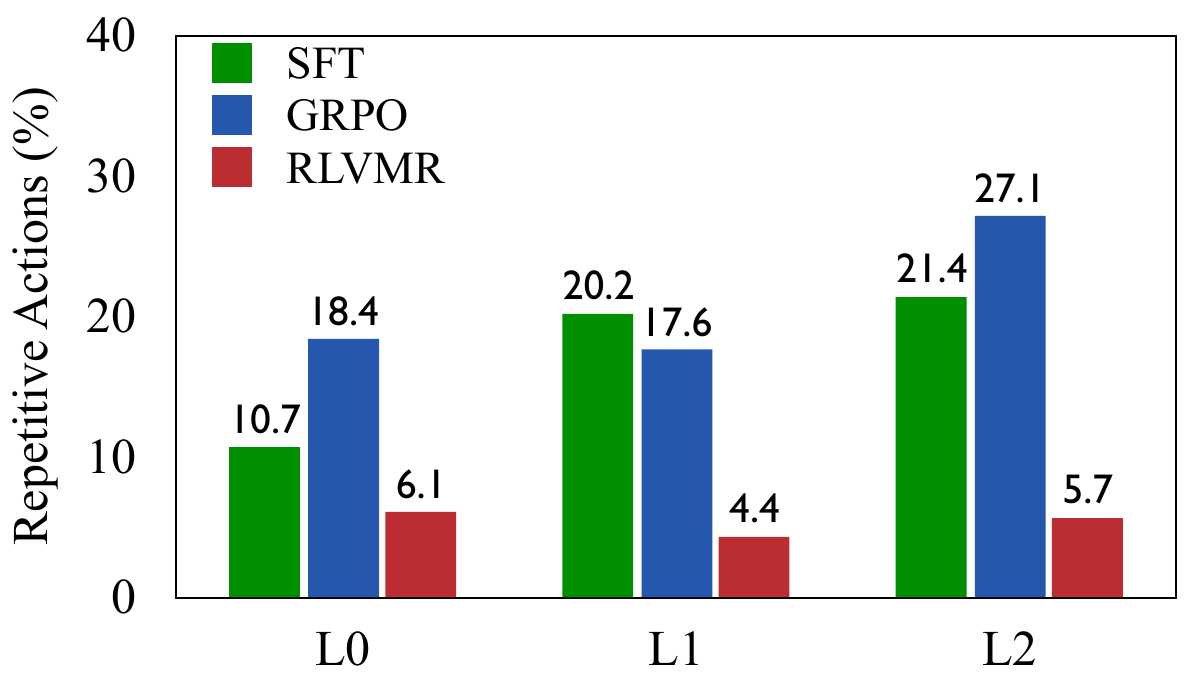}}\\\vspace{5pt}
  \subfloat[Invlaid Actions (7B)]{\includegraphics[width=0.4\linewidth]{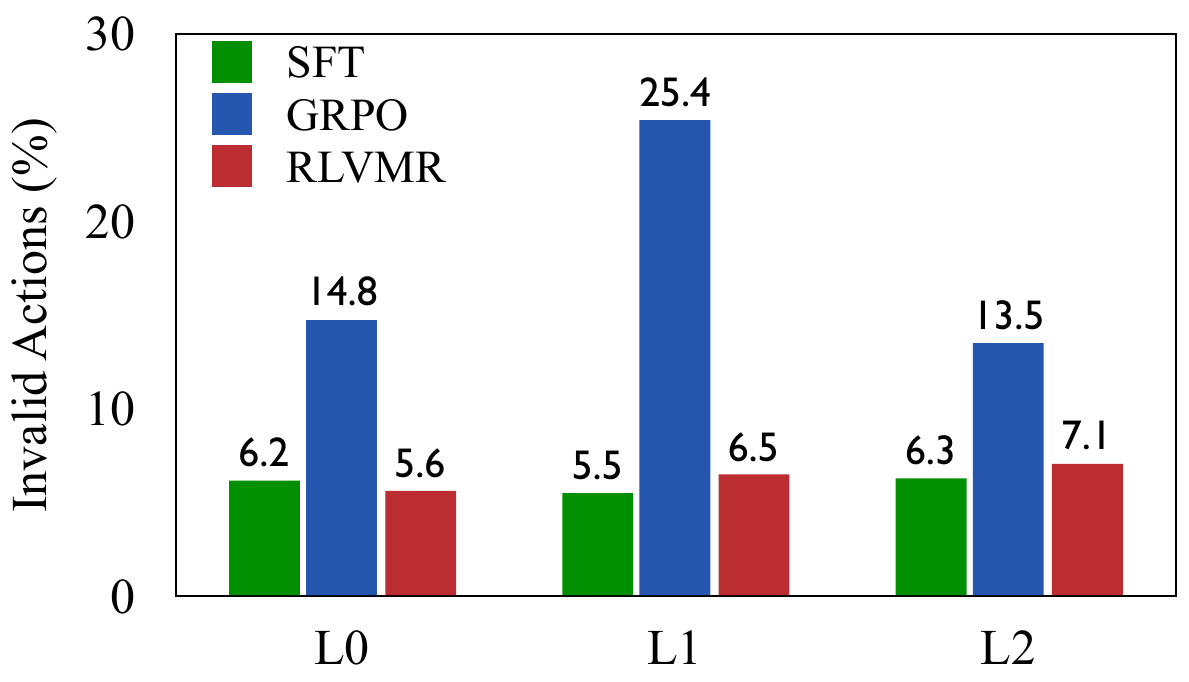}} \hspace{0.1\linewidth}
  \subfloat[Repetitive Actions (7B)]{\includegraphics[width=0.4\linewidth]{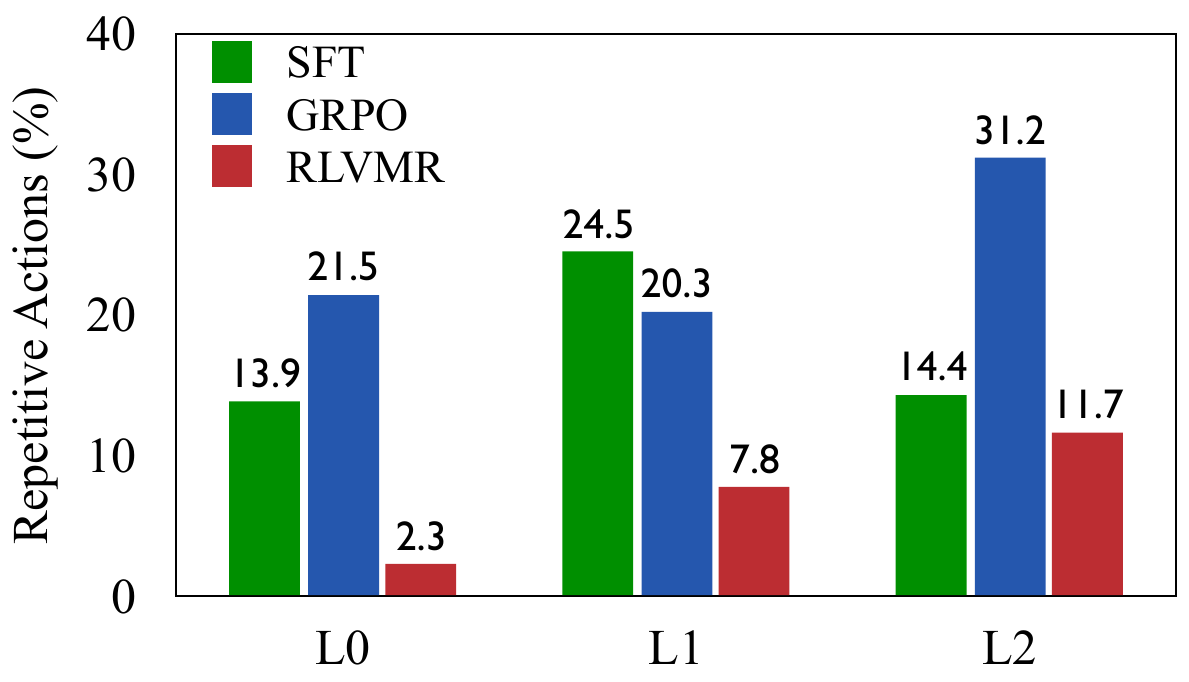}}
  \caption{Exploration efficiency of \method{} compared to SFT and GRPO baselines on ALFWorld. RLVMR consistently and significantly reduces both invalid and repetitive actions across all generalization levels and model sizes, demonstrating its effectiveness at mitigating inefficient exploration.}  \label{fig:exploration-efficiency}
\end{figure}

In this section, we analyze the agent's behavior during task execution to quantify its exploration efficiency, as shown in Figure~\ref{fig:exploration-efficiency}. By comparing RLVMR against strong baselines, we provide direct evidence that our verifiable meta-reasoning rewards successfully cultivate more purposeful and efficient problem-solving strategies.

\paragraph{\method{} directly mitigates the inefficient exploration problem by cultivating highly efficient and purposeful behavior.}
Our approach drastically improves exploration efficiency, providing a direct solution to the problem identified in our contributions. RLVMR significantly reduces both invalid and repetitive actions compared to GRPO, which suffers from inefficient exploration despite its high success rate. For instance, our 7B model reduces the repetitive action rate to just 2.3\% on seen tasks (L0), a nearly tenfold improvement over GRPO (21.5\%). Similarly, the invalid action rate is more than halved (5.6\% vs. 14.8\%). This confirms that our verifiable meta-reasoning rewards—such as the format penalty and the reward for exploring new states—successfully guide the agent away from flawed or redundant steps, leading to more direct and effective problem-solving.

\paragraph{The efficiency gains from RLVMR are robust and generalize to unseen tasks, demonstrating superior reasoning quality.}
A key measure of robustness is whether an agent maintains its efficiency when facing novel challenges. Our analysis confirms that RLVMR excels here. While the GRPO-trained 7B agent's inefficiency worsens on unseen tasks—with its repetitive action rate ballooning from 21.5\% on L0 to 31.2\% on the hardest L2 split—our RLVMR-trained agent remains highly efficient, with its rate staying controlled at 11.7\%. This trend holds for invalid actions as well. This result provides strong evidence for our claim that RLVMR instills a more robust and generalizable reasoning process. Instead of overfitting to familiar paths, our agent learns core problem-solving principles that remain effective in new situations.

\paragraph{Reduced inefficiency serves as strong evidence of enhanced self-correction and error recovery capabilities.}
The marked reduction in flawed behaviors is a direct indicator of the agent's improved ability to self-correct, a cornerstone of robust intelligence. Low rates of repetitive and invalid actions imply that when the agent encounters an error or a dead end, it is less likely to get stuck in a loop~\citep{yuan2025agentr}. This behavior is a direct outcome of our process-level rewards. The `exploration` reward discourages repeating past actions, while the `reflection` reward (as described in our methodology) incentivizes analyzing failures to find a new, corrective path. These results validate our central hypothesis: by explicitly rewarding the {\em process} of good reasoning, RLVMR produces agents that not only find correct answers but also demonstrate the intelligent, adaptive, and robust behavior required to solve complex, long-horizon tasks.

\paragraph{Case Study} Below is a trajectory segment of an agent trained with RLVMR on ALFWorld, performing a novel task: \textit{find two soapbars and put them in the cabinet}. We observe that, guided by meta-cognitive rewards, the agent demonstrates a more comprehensive and meticulous reasoning process, exhibiting significantly more adaptive reasoning patterns. At step 13, the agent is already carrying one soapbar and is actively searching for the second. Lacking explicit information about the soapbars' location, the agent engages in structured and efficient exploration by systematically hypothesizing possible locations via exploratory actions. Concurrently, it keeps track of its current sub-goal, thereby avoiding the task confusion commonly seen in SFT-trained agents.

Notably, in step 14, the agent experiences a brief hallucination, leading to an ineffective action. However, at step 15, rather than falling into an error loop or blindly continuing, the agent immediately triggers reflective behavior, proposing to "double-check my inventory" to collect more information and guard against the possibility of having already acquired both soapbars. This demonstrates a more rigorous and robust reasoning and decision-making process. Experimental results show that such meta-cognitively driven behavioral patterns enhance agent performance and generalization on previously unseen tasks compared to baselines.

\begin{tcolorbox}[title = {A trajectory segment of an agent trained with RLVMR}, breakable]
\textbf{Step 13}

\textbf{Observation:} You arrive at sinkbasin 2. On the sinkbasin 2, you see nothing.\\
\textbf{Reasoning:} \textless explore\textgreater Since there are no soapbars at sinkbasin 2, I should broaden my search to other potential locations. The next logical area to check is the toilet, as soapbars could potentially be found there.\textless/explore\textgreater\\
\textbf{Action:} go to toilet 1\\

\textbf{Step 14}

\textbf{Observation:} You arrive at toilet 1. On the toilet 1, you see a candle 1 and a spraybottle 3.\\
\textbf{Reasoning:} \textless monitor\textgreater My current sub-goal is to obtain the second soapbar. Since there are no soapbars visible here, I will attempt to examine the environment further, just in case the second soapbar is present but not immediately listed in the observation.\textless/monitor\textgreater\\
\textbf{Action:} examine soapbar 2\\

\textbf{Step 15}

\textbf{Observation:} Nothing happens.\\
\textbf{Reasoning:} \textless reflection\textgreater Reflecting on my recent actions, I realize I failed to place the second soapbar in the cabinet earlier. To proceed efficiently, I need to double-check my inventory and ensure I’m actually carrying both soapbars before trying again.\textless/reflection\textgreater\\
\textbf{Action:} inventory\\

\textbf{Step 16}

\textbf{Observation:} You are carrying: a soapbar 1.\\
\textbf{Reasoning:} \textless monitor\textgreater At this point, my sub-goal is still to locate the second soapbar. Since I am currently only carrying soapbar 1, I need to continue searching for the other one.\textless/monitor\textgreater\\
\textbf{Action:} go to countertop 1\\

\textbf{Step 17}

\textbf{Observation:} You arrive at countertop 1. On the countertop 1, you see a soapbottle 1, a spraybottle 1, and a toiletpaper 1.\\
\textbf{Reasoning:} \textless explore\textgreater No soapbar is found on the countertop either, so I should expand my search to nearby areas. Checking places like the toilet paper hanger may help ensure I’m not missing any possible locations for the second soapbar.\textless/explore\textgreater\\
\textbf{Action:} go to toiletpaperhanger 1\\

\textbf{Step 18}

\textbf{Observation:} On the toiletpaperhanger 1, you see a toiletpaper 3.\\
\textbf{Reasoning:} \textless explore\textgreater Since I still haven’t found a new soapbar, I should continue exploring other potential areas, such as the hand towel holders, to make sure I don’t overlook any possible spots where a soapbar could be located.\textless/explore\textgreater\\
\textbf{Action:} go to handtowelholder 1\\
\end{tcolorbox}

\subsection{Training Efficiency}

\begin{table}[t]
\centering
\caption{Comparison of the average lengths of action sequences of different methods across various scenarios. A shorter length (number of actions) indicates higher efficiency.}
\begin{tabular}{ll ccccc}
\toprule
\textbf{Environment} & \textbf{Level}   &   \bf Vanilla &   \bf SFT & \textbf{GRPO} & \textbf{GiGPO} & \textbf{RLVMR} \\
\midrule
\multirow{3}{*}{ALFWorld}    
    & Seen-L0       & 28.8 & 23.2 & 15.8 & 12.6 & \textbf{10.8}\\
    & Unseen-L1     & 29.1 & 24.5 & 18.1 & 14.7 & \textbf{11.6}\\
    & Unseen-L2     & 28.9 & 27.5 & 21.7 & 19.4 & \textbf{15.4}\\
\midrule
\multirow{3}{*}{ScienceWorld}
    & Seen-L0       & 25.8 & 22.9 & 15.4 & 14.7 & \textbf{12.5}\\
    & Unseen-L1     & 27.9 & 25.7 & 26.7 & 25.2 & \textbf{18.8}\\
    & Unseen-L2     & 26.8 & 26.2 & 27.6 & 26.3 & \textbf{20.5}\\
\bottomrule
\end{tabular}
\label{tab:len_comparison}
\end{table}

\begin{figure}[t]
  \centering
    \includegraphics[width=0.8\linewidth]{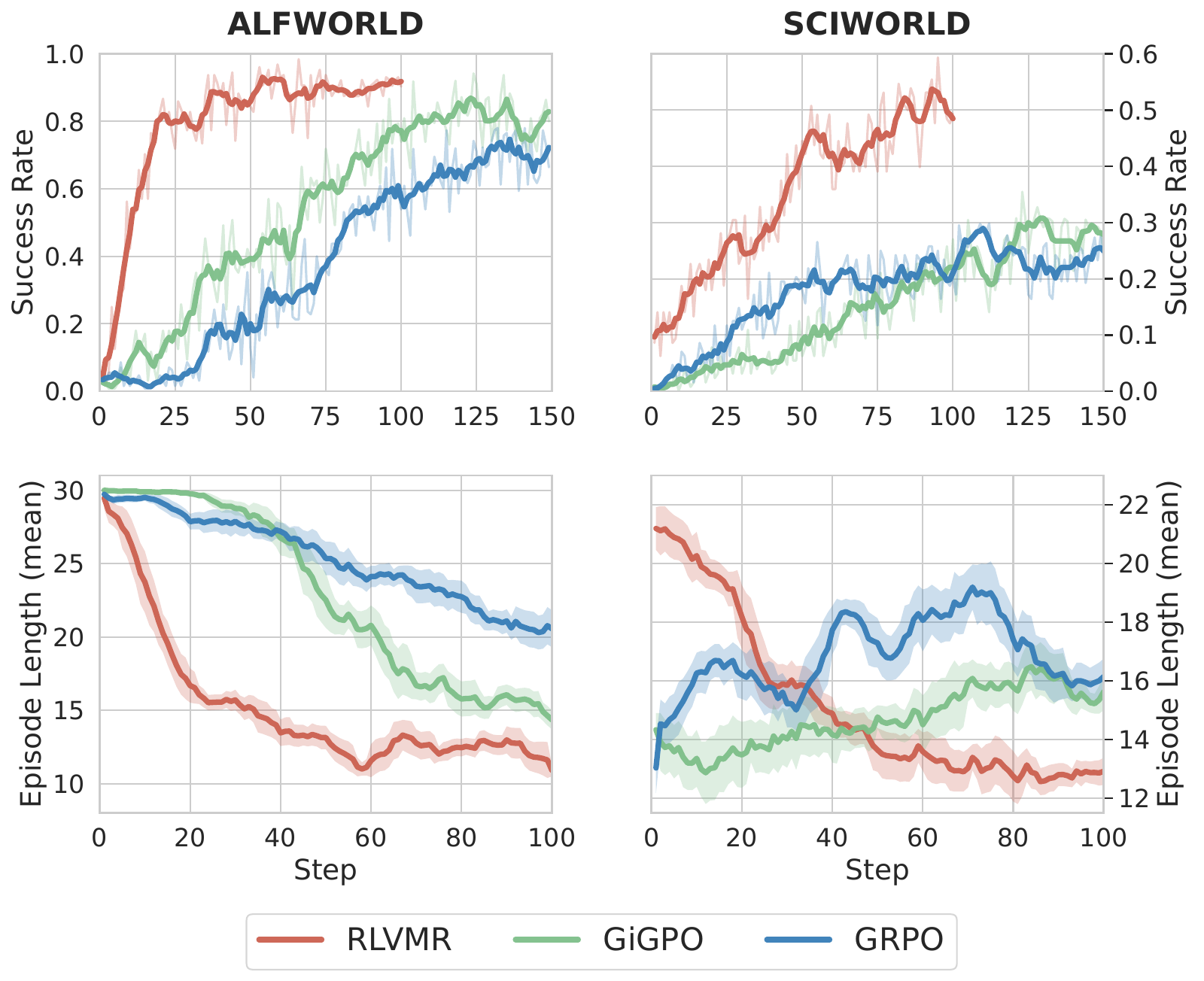}
    \caption{The step count curves of different methods on two datasets during the RL training process.}
    \label{fig:traj_length}
\end{figure}

Beyond achieving higher success rates and more efficient exploration, it is crucial to evaluate how efficiently a method learns. In this section, we analyze the training efficiency of \method{} from two perspectives: the quality of the learned policy, measured by the average length of action trajectories (Table~\ref{tab:len_comparison}), and the stability of the learning process itself (Figure~\ref{fig:traj_length}). We demonstrate that agents trained with \method{} not only find more direct solutions to tasks but also exhibit a more stable and rapid convergence during training compared to baseline methods. This highlights that our process-level rewards provide a clearer and more consistent learning signal, leading to faster and more robust policy optimization.

\paragraph{\method{} learns significantly more efficient policies, reducing the number of actions required to solve tasks.}
As shown in Table~\ref{tab:len_comparison}, agents trained with \method{} consistently find shorter solution paths compared to those trained with GRPO and GiGPO. For instance, on the challenging Unseen-L2 split of ALFWorld, \method{} requires only 15.4 actions on average, a 28.1\% reduction from GRPO (21.7 actions) and a 20.6\% reduction from GiGPO (19.4 actions). This superior efficiency directly addresses the ``inefficient exploration issue'' by penalizing redundant or invalid actions. We attribute this improvement to the verifiable meta-reasoning rewards: the exploration reward discourages revisiting states, while the reflection reward encourages escaping unproductive loops. This mechanism cultivates more direct and purposeful reasoning, enhancing agent robustness as claimed in our contributions.

\paragraph{\method{} demonstrates superior training stability and faster convergence to efficient solutions.}
Figure~\ref{fig:traj_length} illustrates the training dynamics, showing that \method{} not only achieves shorter final trajectory lengths but also converges much faster. While baseline methods (e.g., GRPO and GiGPO) exhibit unstable or even increasing action counts during training, \method{}'s action count curve shows a stable and consistent decline. This is particularly evident on ScienceWorld, where baselines struggle. Their action counts increase as they learn to attempt more complex actions without a clear reasoning strategy. In contrast, \method{}'s cold-start phase provides foundational knowledge, and its dense meta-reasoning rewards offer a stable, process-level learning signal. This prevents the agent from developing inefficient, looping behaviors and validates that our framework cultivates robust reasoning processes rather than just optimizing for final outcomes.

\subsection{Ablation Study}
\label{sec:ablation}

\begin{table}[t]
\centering
\caption{Ablation results on ALFWorld and ScienceWorld (success rates (\%) on L2 variant).}
\begin{tabular}{lrr}
\toprule
\textbf{Variant} & \textbf{ALFWorld} & \textbf{ScienceWorld} \\
\midrule
\method{} (Full) & \textbf{56.3} & \textbf{26.5} \\
\quad w/o $A^{\text{T}}$ (Outcome Reward) & 12.5 & 7.8 \\
\quad w/o $A^{\text{MC}}$ (Meta-Reasoning Reward) & 45.3 & 20.3 \\
\quad w/o CS (Cold-Start) & 40.6 & 18.8 \\
\bottomrule
\end{tabular}
\label{tab:ablation}
\end{table}

We conduct ablation studies on the Unseen-L2 split using Qwen2.5-1.5B-Instruct to analyze the impact of our framework's key components: the trajectory-level outcome advantage signal ($A^{\text{T}}$), the meta-reasoning advantage signal ($A^{\text{MC}}$), and the cold-start process (CS). The results in Table~\ref{tab:ablation} confirm that each component is critical for achieving optimal performance.

\paragraph{Verifiable meta-reasoning rewards are essential for tackling complex, unseen tasks.}
Removing the meta-reasoning advantage signal ($A^{\text{MC}}$) causes a significant performance drop, with the success rate on ALFWorld falling by 11.0 percentage points (from 56.3\% to 45.3\%) and on ScienceWorld by 6.2 points. This variant is equivalent to a standard GRPO agent fine-tuned from the cold-start model. The sharp decline validates our central hypothesis: directly rewarding beneficial reasoning processes is crucial for developing robust problem-solving skills. This component directly addresses the ``inefficient exploration issue'' by providing dense, process-level signals that guide the agent toward more efficient and logical behaviors, a benefit that outcome-only rewards ($A^{\text{T}}$) cannot provide alone.

\paragraph{Outcome-based rewards remain indispensable for guiding the agent toward final task success.}
Eliminating the trajectory-level outcome advantage ($A^{\text{T}}$) results in a catastrophic performance collapse, with the success rate plummeting to just 12.5\% on ALFWorld and 7.8\% on ScienceWorld. This demonstrates that while meta-reasoning rewards effectively shape the {\bf process}, the global signal of task success is vital for orienting the agent toward the ultimate goal. The meta-reasoning rewards are locally effective—for instance, rewarding non-repetitive exploration—but without the final outcome signal, the agent cannot learn which explorations ultimately lead to a successful trajectory. This confirms that the synergy between process-level and outcome-level rewards is a key strength of the \method{} framework.

\paragraph{A lightweight cold-start phase is critical for bootstrapping the agent's reasoning capabilities.}
Training the agent without the supervised fine-tuning cold-start (CS) phase leads to a substantial performance decrease on both ALFWorld (down 15.7 points) and ScienceWorld (down 7.7 points). The cold-start phase, which uses only 200 trajectories, is not intended to solve the tasks but to equip the model with the basic ability to generate syntactically correct meta-reasoning tags and follow instructions. For smaller models (e.g., 1.5B), this initial grounding is vital; without it, the agent often fails to produce parsable outputs during RL, leading to training instability and policy collapse. This finding underscores the efficiency of our approach: a brief, low-data cold-start is sufficient to unlock the model's capacity for complex reasoning, which is then honed by the RL phase.

\section{Related Work}

\paragraph{LLM Reinforcement Learning}

Reinforcement learning (RL) has been instrumental in aligning large language models (LLMs) with human preferences. Prominent examples include Reinforcement Learning from Human Feedback (RLHF)~\citep{ouyang2022RLHF} and Direct Preference Optimization (DPO)~\citep{rafailov2023DPO}. 
Beyond alignment, recent work has also leveraged RL to enhance other crucial LLM capabilities, such as reasoning~\citep{hu2025open, muennighoff2025s1} and emotional intelligence~\citep{rlver}.
Recently, group-based RL algorithms have emerged as a promising alternative, with methods like GRPO~\citep{feng2025GRPO}, Dr.GRPO~\citep{liu2025DrGRPO}, and DAPO~\citep{yu2025DAPO} estimating advantages by using batches of samples generated from the same prompt. In contrast to actor-critic methods like PPO~\citep{schulman2017PPO}, this approach to advantage estimation does not require an additional critic model, making large-scale RL training for LLMs more computationally efficient and practical. These approaches have demonstrated significant effectiveness in tasks such as mathematical reasoning, search, and tool use~\citep{yu2025DAPO, hu2025open}. However, applying these RL methods to multi-turn, long-horizon tasks remains a significant challenge, primarily due to issues of sparse and delayed rewards~\citep{wang2025ragen}, which is the focus of our work.

\paragraph{LLM Agents}
Large language models (LLMs) are now widely utilized as the core of agentic systems across diverse domains, including code generation~\citep{huang2023agentcoder, zhang2024codeagent}, web interaction~\citep{bai2024digirl, agashe2024agents, abuelsaad2024agente}, embodied intelligence~\citep{zeng2024agenttuning, qiao2024WKM, fu2025agentrefine}, and emotional intelligence~\citep{sage}. Early approaches primarily leveraged existing pretrained models, employing sophisticated prompting strategies and external tools to enhance performance on complex tasks, as exemplified by methods like ReAct~\citep{yao2023react, shinn2023reflexion}. However, models with smaller parameter counts often lack the requisite foundational capabilities for such complex reasoning. To address this limitation, some studies employ supervised fine-tuning (SFT) to enhance the models' decision-making abilities~\citep{zhang2024you, xi2024agentgym, qin2024toolllm}.
Other approaches explore single-step or offline reinforcement learning to further augment agent performance~\citep{yu2024steptool,xiong2024watch,zhou2024archer}. More recently, there has been a growing focus on end-to-end reinforcement learning for agents~\citep{wang2025ragen, feng2025gigpo}, which learn through direct, adaptive online interaction with an environment, thereby obviating the need for complex data preparation or step-level reward models. Nevertheless, these approaches still grapple with challenges in fine-grained credit assignment and generalization~\citep{wang2025ragen}. In this work, we introduce a novel approach to reward shaping that draws from the perspective of verifiable meta-cognitive behaviors. By explicitly encouraging effective reasoning patterns, our method enhances agent performance and robustness.

\section{Conclusion}

In this work, we addressed the critical problem of {\bf inefficient exploration} in long-horizon agents, where standard RL rewards successful outcomes without ensuring the coherence of the underlying reasoning process. We introduced {\bf \method{}}, a novel framework that integrates process-level supervision by providing dense, verifiable rewards for explicit meta-reasoning behaviors such as planning, exploration, and reflection. Our approach, which combines a lightweight cold-start phase with end-to-end policy optimization, effectively shapes the agent's reasoning process to be more robust, efficient, and adaptive.

Extensive experiments on ALFWorld and ScienceWorld demonstrate that \method{} establishes a new state of the art, significantly improving success rates and generalization to unseen tasks. Crucially, we showed that these performance gains are driven by tangible improvements in reasoning quality, including fewer redundant actions and a markedly improved ability to recover from errors. This work underscores the importance of supervising the reasoning process itself and offers a scalable, effective method for building more reliable and generalizable autonomous agents. Future work could extend \method{} to multi-modal domains, explore more sophisticated adaptive reward mechanisms, and apply the framework to complex, real-world scenarios such as robotics and software engineering, advancing the development of more capable and trustworthy AI systems.

\bibliography{main}
\bibliographystyle{colm2024_conference}

\clearpage

\clearpage
\appendix

\section{Setup Details}
\label{app:setup}

\subsection{Dataset Details}
\label{app:dataset}

\textbf{ALFWorld} is a household task environment that requires agents to explore rooms and employ common-sense reasoning to accomplish tasks, such as "put the pencil on the desk." The environment provides feedback on whether the agent successfully completes the task within a given number of steps. ALFWorld describes the environment in purely textual form and supplies a reward signal indicating only the final task completion status.

\textbf{ScienceWorld} is a text-based virtual environment designed as a comprehensive testbed for evaluating and enhancing scientific reasoning abilities in AI systems. It features tasks spanning 10 scientific domains and 30 subcategories, simulating a broad range of experiments found in elementary science curricula, including state changes of matter, measurement, electricity, life sciences, plant growth, chemical reactions, genetics, among others. Each task involves multiple subgoals, and the final reward is computed based on the completion of these subgoals. However, to better reflect real-world scenarios, we only use the final reward and disregard intermediate rewards. Notably, some tasks in ScienceWorld require agents to make conclusive judgments based on experimental outcomes or common sense; a task is considered successful only if the agent provides the correct final answer.

Both ALFWorld and ScienceWorld offer ``seen'' and ``unseen'' variants for evaluating generalization capabilities. To further assess the agents’ robustness and generalization, we define three difficulty levels (L0, L1, L2), with L2 comprising entirely held-out task types. Specifically, for ALFWorld, we designate \textit{Cool \& Place} and \textit{Pick Two \& Place} as held-out tasks; for ScienceWorld, the final task type of each topic is reserved for unseen evaluation.

\subsection{Implementation Details}\label{app:implement-details}

We conducted experiments on both the Qwen2.5-1.5B-Instruct and Qwen2.5-7B-Instruct models. During the cold start phase, we set the batch size per GPU to 16, used a learning rate of $1 \times 10^{-5}$, and trained for 5 epochs. For the RL phase, we adopted the veRL framework with necessary modifications. The batch size per GPU was also set to 16. At each training step, we sampled from 16 distinct environments, with each environment rolling out 8 trajectories. The weights for outcome advantage and meta-reasoning advantage were both set to 0.5 by default. To penalize outputs that did not adhere to the required format, we applied a reward penalty of -0.1. Specifically, an output was considered valid only if it included at least one meta-reasoning tag (e.g., $\langle \text{reflection} \rangle$) and one action tag (e.g., $\langle \text{action} \rangle$). The KL regularization coefficient was set to 0.01. For all environments, we allowed a maximum of 30 steps per episode by default.

\section{Detailed Experiment Results}\label{app:detailed-results}
We further report the success rates of different methods on various tasks in ALFWorld. Table~\ref{tab:alfworld-1.5B} provides the results using the Qwen2.5-1.5B model as the base model, while Table~\ref{tab:alfworld-7B} presents the results using the Qwen2.5-7B model. As shown in the tables, RLVMR generally outperforms other methods across all tasks, and particularly exhibits strong performance in more complex tasks. This demonstrates that RLVMR, by rewarding high-quality reasoning behaviors, significantly enhances the robustness and adaptability of agents in multi-step interactions.

\begin{table*}[h]
\centering
\begin{tabular}{lccccccc}
\toprule
\textbf{Method} & \textbf{Pick} & \textbf{Look} & \textbf{Clean} & \textbf{Heat} & \textbf{Cool} & \textbf{Pick2} & \textbf{All} \\
\midrule
ReAct   & 23.1 & 18.3 & 10.8 &  8.7 &  3.5 &  0.0 & 13.7 \\
+SFT    & 43.2 & 42.0 & 35.9 & 33.2 & 29.4 & 29.7 & 38.7 \\
+ETO    & 73.6 & 46.3 & 66.2 & 68.3 & 62.8 & 55.6 & 66.4 \\
+GRPO   & 80.3 & 55.6 & 88.1 & 76.2 & 62.0 & 72.1 & 71.1 \\
+GiGPO  & 92.8 & 66.5 & 90.7 & 90.9 & 80.2 & 73.8 & 83.2 \\
+RLVMR  & 95.2 & 78.8 & 91.2 & 90.2 & 83.9 & 77.6 & 87.9 \\
\bottomrule
\end{tabular}
\caption{Success rates on ALFWorld with Qwen2.5-1.5B model.}
\label{tab:alfworld-1.5B}
\end{table*}

\begin{table*}[h]
\centering
\begin{tabular}{lccccccc}
\toprule
\textbf{Method} & \textbf{Pick} & \textbf{Look} & \textbf{Clean} & \textbf{Heat} & \textbf{Cool} & \textbf{Pick2} & \textbf{All} \\
\midrule
ReAct   & 43.1 & 33.2 & 18.7 & 16.4 & 20.2 & 12.8 & 28.5 \\
+SFT    & 70.8 & 63.0 & 61.1 & 46.3 & 49.7 & 33.2 & 57.0 \\
+ETO    & 88.2 & 70.5 & 82.3 & 83.6 & 71.0 & 51.2 & 74.2 \\
+GRPO   & 90.2 & 76.7 & 86.0 & 80.1 & 68.3 & 56.4 & 77.3 \\
+GiGPO  & 91.7 & 85.9 & 93.3 & 90.3 & 89.0 & 83.6 & 90.2 \\
+RLVMR  & 95.3 & 88.2 & 90.1 & 92.4 & 89.8 & 86.7 & 91.8 \\
\bottomrule
\end{tabular}
\caption{Success rates on ALFWorld with Qwen2.5-7B model.}
\label{tab:alfworld-7B}
\end{table*}

\section{Prompts}\label{app:prompts}

Below are the prompts we used in the ALFWorld and ScienceWorld environments.


\begin{tcolorbox}[title = {Prompt Template for ALFWorld Enviroment}, breakable]
You are an expert agent operating in the \textbf{ALFRED Embodied Environment}. Your task is to: \texttt{\{task\_description\}}

\vspace{0.2em}

\textbf{Prior to this step, you have already taken \texttt{\{step\_count\}} step(s).}

Below are the most recent \texttt{\{history\_length\}} observations and the corresponding actions you took: \texttt{\{action\_history\}}

You are now at step \texttt{\{current\_step\}} and your current observation is: \texttt{\{current\_observation\}}

\vspace{0.5em}

\textbf{Your admissible actions of the current situation are:} \texttt{\{admissible\_actions\}}.

\vspace{0.5em}

\textbf{Your previous overall plan is:} \texttt{\{planning\}}. Please strictly adhere to your plan.

\vspace{0.8em}

Now it's your turn to take an action, following these steps:

\begin{enumerate}
    \item \textbf{First, reason using \emph{ONLY ONE} tag pair and express your reasoning in \emph{one concise, brief sentence}:}
    
    \begin{itemize}
      \item \texttt{<planning>} Plan or replan the entire task by breaking it down into high-level steps. Focus on outlining the full sequence required to complete the overall task, not just the immediate next action. Use this at the beginning of complex tasks or whenever the previous plan is incorrect or insufficient. It is necessary to list all the points separately. eg, step 1: xxx, step 2: xxx, step 3: xxx, etc.
      \item \texttt{<explore>} When results are unexpected or information is lacking, use current observations to think outside the box and list as many possible locations, items, or actions as possible. Use this approach when facing obstacles that require creative and innovative thinking.
      \item \texttt{<reflection>} Analyze the reasons for errors in task execution and correct them by exploring alternative approaches. 'No known action matches that input.' indicates the action is invalid. This is typically used when several consecutive actions yield no substantial progress.
      \item \texttt{<monitor>} Continuously track the current progress and history of reasoning and execution throughout the task. Recall the current subgoal and consider the next concrete action, ensuring agent alignment with the overall plan. Typically used when task outcomes are as expected and no other mode of reasoning is required.
    \end{itemize}

    \item \textbf{After your reasoning, you \emph{MUST} select and present an admissible action for the current step within \texttt{<action>} \ldots \texttt{</action>} tags.}

    Specify the next action the agent should take to progress toward the task goal, following these guidelines:
    \begin{enumerate}
        \item \textbf{Object and Receptacle References:} Use specific identifiers:
        \begin{itemize}
            \item \texttt{[obj id]} for objects (e.g., apple 1).
            \item \texttt{[recep id]} for receptacles (e.g., countertop 1).
        \end{itemize}
        \item \textbf{Action Validity:} Follow the exact format below. Any deviation renders the action invalid:
        \begin{itemize}
            \item Valid actions: \texttt{go to [recep id]}, \texttt{take [obj id] from [recep id]}, \texttt{put [obj id] in/on [recep id]}, \texttt{open/close [recep id]}, \texttt{use [obj id]}, \texttt{heat/cool/clean [obj id] with [recep id]}.
        \end{itemize}
    \end{enumerate}
\end{enumerate}

\end{tcolorbox}

\begin{tcolorbox}[title = {Prompt Template for ScienceWorld Environment}, breakable]
You are an expert agent operating in the \textbf{ScienceWorld} environment, which is a text-based virtual environment centered around accomplishing tasks from the elementary science curriculum.

\vspace{0.2em}

\textbf{Your current task is:} \texttt{\{task\_description\}}

\vspace{0.5em}

\textbf{Prior to this step, you have already taken \texttt{\{step\_count\}} step(s).} 

Below are the most recent \texttt{\{history\_length\}} observations and the corresponding actions you took: \texttt{\{action\_history\}}

You are now at step \texttt{\{current\_step\}} and your current observation is: \texttt{\{current\_observation\}}

\vspace{0.5em}

\textbf{Here are the actions you may take:}
\begin{itemize}
    \item \texttt{\{"action": "open OBJ", "description": "open a container"\}}
    \item \texttt{\{"action": "close OBJ", "description": "close a container"\}}
    \item \texttt{\{"action": "activate OBJ", "description": "activate a device"\}}
    \item \texttt{\{"action": "deactivate OBJ", "description": "deactivate a device"\}}
    \item \texttt{\{"action": "connect OBJ to OBJ", "description": "connect electrical components"\}}
    \item \texttt{\{"action": "disconnect OBJ", "description": "disconnect electrical components"\}}
    \item \texttt{\{"action": "use OBJ [on OBJ]", "description": "use a device/item"\}}
    \item \texttt{\{"action": "look around", "description": "describe the current room"\}}
    \item \texttt{\{"action": "look at OBJ", "description": "describe an object in detail"\}}
    \item \texttt{\{"action": "look in OBJ", "description": "describe a container's contents"\}}
    \item \texttt{\{"action": "read OBJ", "description": "read a note or book"\}}
    \item \texttt{\{"action": "move OBJ to OBJ", "description": "move an object to a container"\}}
    \item \texttt{\{"action": "pick up OBJ", "description": "move an object to the inventory"\}}
    \item \texttt{\{"action": "put down OBJ", "description": "drop an inventory item"\}}
    \item \texttt{\{"action": "pour OBJ into OBJ", "description": "pour a liquid into a container"\}}
    \item \texttt{\{"action": "dunk OBJ into OBJ", "description": "dunk a container into a liquid"\}}
    \item \texttt{\{"action": "mix OBJ", "description": "chemically mix a container"\}}
    \item \texttt{\{"action": "go to LOC", "description": "move to a new location"\}}
    \item \texttt{\{"action": "eat OBJ", "description": "eat a food"\}}
    \item \texttt{\{"action": "flush OBJ", "description": "flush a toilet"\}}
    \item \texttt{\{"action": "focus on OBJ", "description": "signal intent on a task object"\}}
    \item \texttt{\{"action": "wait", "description": "take no action for 10 iterations"\}}
    \item \texttt{\{"action": "wait1", "description": "take no action for 1 iteration"\}}
    \item \texttt{\{"action": "task", "description": "describe current task"\}}
    \item \texttt{\{"action": "inventory", "description": "list your inventory"\}}
\end{itemize}

\vspace{0.5em}

\textbf{Your previous overall plan is:} \texttt{\{planning\}}.

Please strictly adhere to your plan.

\vspace{0.8em}

Now it's your turn to take an action, following these steps:

\begin{enumerate}
    \item \textbf{First, reason using \emph{ONLY ONE} tag pair and express your reasoning in \emph{one concise, brief sentence}:}
    
    \begin{itemize}
      \item \texttt{<planning>} \\
        Plan or replan the entire task by breaking it down into high-level steps. Focus on outlining the full sequence required to complete the overall task, not just the immediate next action. \\
        Use this at the beginning of complex tasks or whenever the previous plan is incorrect or insufficient. \\
        It is necessary to list all the points separately. eg, step 1: xxx, step 2: xxx, step 3: xxx, etc.
      \item \texttt{<explore>} \\
        When results are unexpected or information is lacking, use current observations to think outside the box and list as many possible locations, items, or actions as possible. \\
        Use this approach when facing obstacles that require creative and innovative thinking.
      \item \texttt{<reflection>} \\
        Analyze the reasons for errors in task execution and correct them by exploring alternative approaches. 'No known action matches that input.' indicates the action is invalid. \\
        This is typically used when several consecutive actions yield no substantial progress.
      \item \texttt{<monitor>} \\
        Continuously track the current progress and history of reasoning and execution throughout the task. Recall the current subgoal and consider the next concrete action, ensuring agent alignment with the overall plan. \\
        Typically used when task outcomes are as expected and no other mode of reasoning is required.
    \end{itemize}

    \item \textbf{After your reasoning, you \emph{MUST} select and present an appropriate action for the current step within \texttt{<action>} \ldots \texttt{</action>} tags.}
\end{enumerate}

\end{tcolorbox}

\end{document}